%% file: main.tex
\documentclass[journal]{IEEEtran}
\usepackage{amsmath,amsfonts}
\usepackage{algorithm}
\usepackage{algpseudocode}
\usepackage{float}
\usepackage{array}
\usepackage{textcomp}
\usepackage{stfloats}
\usepackage{url}
\usepackage{verbatim}
\usepackage{graphicx}
\usepackage{cite}
\usepackage{color}
\usepackage{multirow}
\usepackage{subfigure}
\usepackage{bm}
\usepackage{amssymb}
\usepackage{pifont}
\usepackage{booktabs}
\usepackage{ragged2e}
\usepackage{subfigure}
\usepackage{hyperref}
\definecolor{new_yellow}{RGB}{255,209,0}
\definecolor{new_green}{RGB}{105,217,39}

\begin{document}

\title{FocusTrack: A Self-Adaptive Local Sampling Algorithm for Efficient Anti-UAV Tracking}

% authors information
\author{Ying Wang, Tingfa Xu$^{\ast}$ and Jianan Li$^{\ast}$ 
\thanks{*Corresponding author}
\IEEEcompsocitemizethanks{\IEEEcompsocthanksitem 
Ying wang, Tingfa Xu and Jianan Li are with Beijing Institute of Technology, Beijing 100081, China (e-mail: \{3120215325, ciom\_xtf1, lijianan \}@bit.edu.cn).
\IEEEcompsocthanksitem  
Tingfa Xu and Jianan Li are also with the Key Laboratory of Photoelectronic Imaging Technology and System, Ministry of Education of China, Beijing 100081, China 
\IEEEcompsocthanksitem 
Tingfa Xu is also with Chongqing Innovation Center, Beijing Institute of Technology, Chongqing 401135, China.
}
% <-this % stops a space
\thanks{Manuscript received April 19, 2021; revised August 16, 2021.}}

\markboth{Journal of \LaTeX\ Class Files,~Vol.~14, No.~8, August~2021}%
{Shell \MakeLowercase{\textit{et al.}}: A Sample Article Using IEEEtran.cls for IEEE Journals}



\maketitle

\begin{abstract}
Anti-UAV tracking poses significant challenges, including small target sizes, abrupt camera motion, and cluttered infrared backgrounds. Existing tracking paradigms can be broadly categorized into \emph{global-based} and \emph{local-based} methods. Global-based trackers, such as SiamDT\cite{antiuav410} and SiamSTA\cite{siamsta}, achieve high accuracy by scanning the entire field of view but suffer from excessive computational overhead, limiting real-world deployment. In contrast, local-based methods, including OSTrack~\cite{ostrack} and ROMTrack~\cite{romtrack}, efficiently restrict the search region but struggle when targets undergo significant displacements due to abrupt camera motion. 
Through preliminary experiments, it is evident that a local tracker, when paired with adaptive search region adjustment, can significantly enhance tracking accuracy, narrowing the gap between local and global trackers. 
To address this challenge, we propose FocusTrack, a novel framework that dynamically refines the search region and strengthens feature representations, achieving an optimal balance between computational efficiency and tracking accuracy. Specifically, our Search Region Adjustment (SRA) strategy estimates the target presence probability and adaptively adjusts the field of view, ensuring the target remains within focus. 
Furthermore, to counteract feature degradation caused by varying search regions, the Attention-to-Mask (ATM) module is proposed. This module integrates hierarchical information, enriching the target representations with fine-grained details.
Experimental results demonstrate that FocusTrack achieves state-of-the-art performance, obtaining 67.7\% AUC on AntiUAV\cite{antiuav310} and 62.8\% AUC on AntiUAV410\cite{antiuav410}, outperforming the baseline tracker by 8.5\% and 9.1\% AUC, respectively. 
In terms of efficiency, FocusTrack surpasses global-based trackers, requiring only 30G MACs and achieving 143 fps with FocusTrack (SRA) and 44 fps with the full version, both enabling real-time tracking.
Code and models are available at
\url{https://github.com/vero1925/FocusTrack}.

\end{abstract}


\section{Introduction}
\label{introduction}

\IEEEPARstart{T}{h}e rapid growth of the low-altitude economy has driven widespread UAV adoption across various sectors, enhancing efficiency in military, security, and urban management. However, their stealth and agility pose security risks, making the advancement of Anti-UAV tracking technologies essential\cite{smalltrack, zhang2024efficient}.

\begin{figure}[t]
\centering
\includegraphics[scale=0.58]{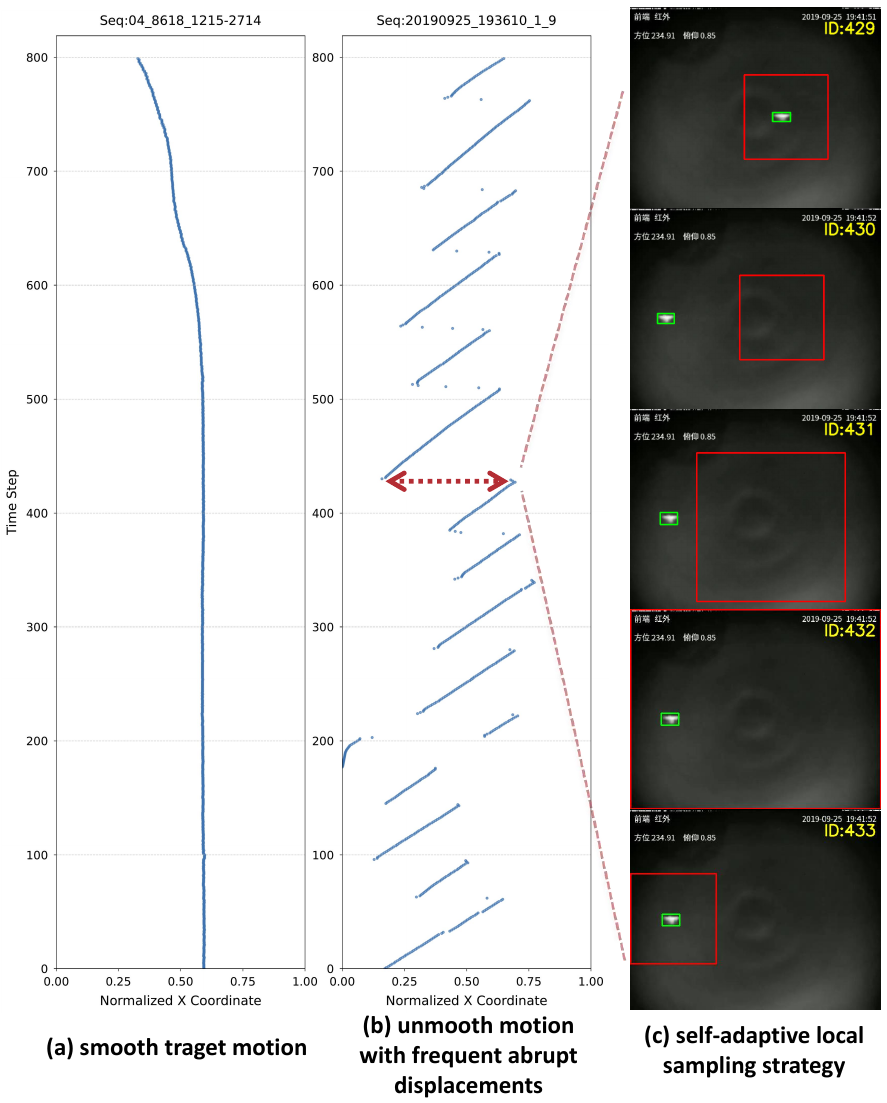} 
\caption
{Illustration of camera motion challenge in Anti-UAV tracking. (a) and (b) show the target’s horizontal position over time in two AntiUAV410\cite{antiuav410} sequences: (a) with smooth motion and (b) with abrupt displacements. (c) visualizes a displacement at frame 430 in (b), where the green box marks the target and the red box the search region. From frames 431–433, FocusTrack adaptively adjusts the region to successfully reacquire the target.}
\label{fig:motivation} 
\end{figure}

As a specialized branch of Single Object Tracking (SOT), Anti-UAV tracking inherits the fundamental task of identifying and tracking an initialized target within a video sequence, with a specific focus on UAV targets. While this specialization simplifies category recognition, it introduces unique challenges beyond conventional SOT:
(1) Thermal Infrared (TIR) Modality: To ensure all-weather operation, Anti-UAV tracking often relies on TIR data, which lacks the rich color and texture features of RGB images, reducing target discriminability. Additionally, thermal images suffer from low resolution, and UAVs frequently blend into the background, making tracking more challenging.
(2) Sensitivity to Camera Motion: The small target size and long-range detection amplify the impact of camera movements, causing significant target displacement within frames and degrading tracking performance. 
Fig.~\ref{fig:motivation} illustrates this effect with two video sequences showing the target's horizontal position over time. In (a), the target exhibits smooth positional changes, whereas (b) reveals frequent abrupt displacements. Further visualization in (c) highlights a drastic shift occurring at frame 430, where the target, initially centered in frame 429, suddenly moves to the left edge in the next frame due to the camera motion.

Experimental results in \cite{antiuav410} indicate that state-of-the-art SOT algorithms\cite{ostrack, mixformer, droptrack} struggle with Anti-UAV tracking, primarily due to their reliance on \emph{local search strategies}. 
Local trackers typically employ a crop-and-resize approach, where the search region is determined by a fixed search factor relative to the target size.
As shown in frame 429 of Fig.~\ref{fig:motivation}(c) , the red bounding box represents the search region of a typical local tracker. When a significant displacement occurs at frame 430, the target moves entirely out of the search region. Without a timely search region adjustment strategy, even the strongest feature extraction networks fail to relocate the target, ultimately leading to tracking failure.

To address this, existing high-precision Anti-UAV tracking algorithms often adopt a global search strategy based on re-detection\cite{siamsta, antiuav410, gasiam, 9607440}, where the entire frame serves as the search region. While this approach effectively handles discontinuous target position changes, its high computational complexity and resource demands severely limit practical deployment.
In contrast, local tracking approaches provide a more efficient alternative by restricting the search region to a limited spatial area, significantly reducing computational overhead. Moreover, by naturally excluding similar-looking objects outside the search region, they mitigate performance degradation caused by distractors.

To fully harness these advantages while maintaining high accuracy, our goal is to bridge the gap between local and global trackers, enabling Anti-UAV tasks to benefit from the efficiency of local tracking while achieving performance comparable to global methods. And the key is to properly adjust the search region.

\begin{figure}[t]
\centering
\includegraphics[scale=0.22]{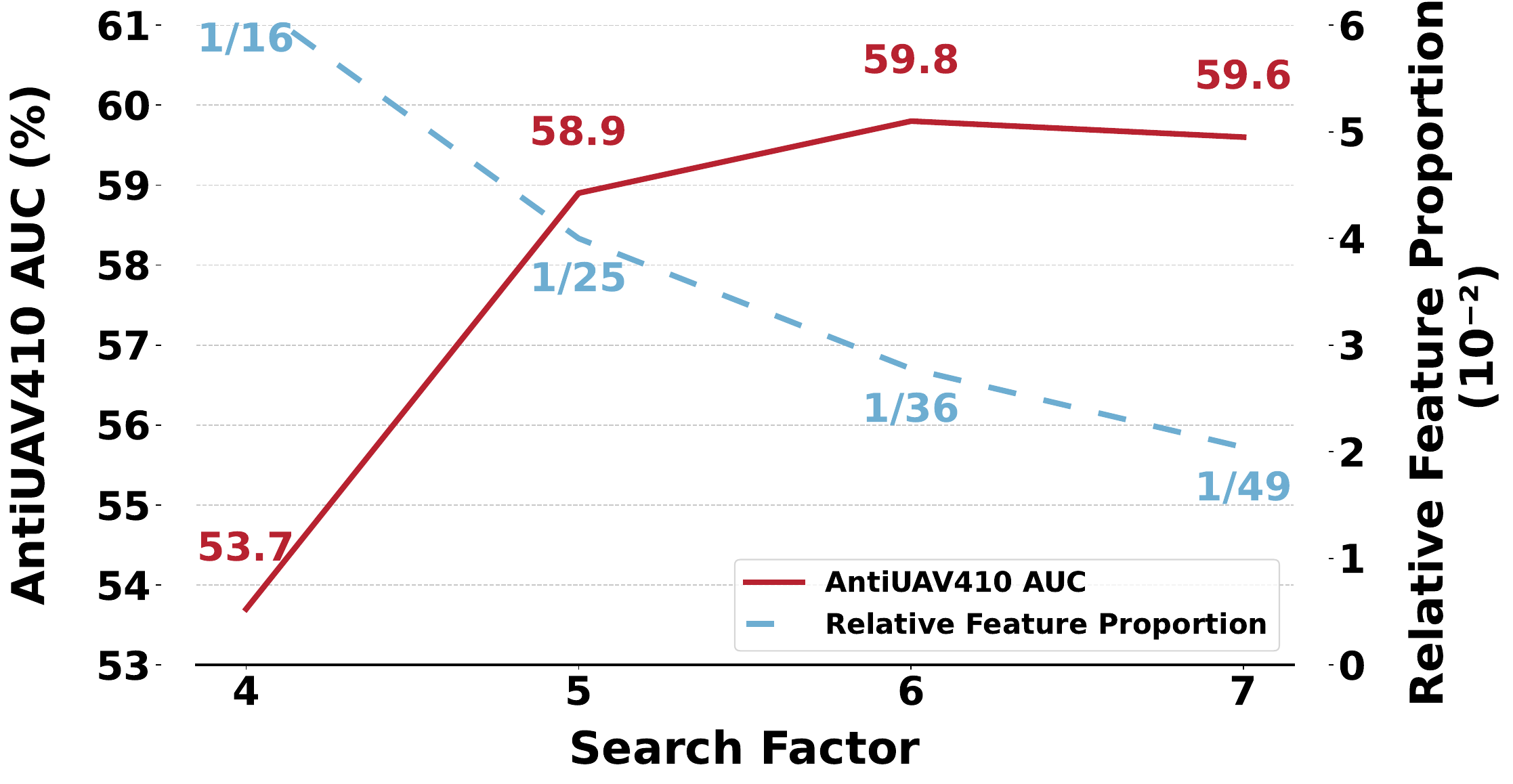}
\caption{
Preliminary experiments on OSTrack~\cite{ostrack} to explore the impact of search factors. The red solid line shows tracking performance (AUC) on AntiUAV410, while the blue dashed line represents the relative feature proportion.
} 
\label{fig:experiment} 
\end{figure}

Prior experiments have shown that expanding the search region can significantly improve tracking accuracy by analyzing its impact on performance.
As illustrated in Figure~\ref{fig:experiment}, expanding the search factor from 4 to 6 led to an improvement of 6.1\% AUC (from 53.7\% to 59.8\%), demonstrating the benefit of a larger search area. However, further increasing the search factor to 7 resulted in performance degradation. This decline can be attributed to two key factors: (1) A larger search region reduces the target’s relative feature proportion, leading to weaker target representations. For example, with a search factor of 7, the target occupies only $1/49$ of the total feature space, compared to $1/16$ with a factor of 4. (2) Expanding the field of view introduces additional background clutter, increasing the difficulty of distinguishing the target from distractors.

These observations suggest that the search region should ideally be dynamically adapted to the scene dynamics rather than being fixed. 
Based on this insight, an adaptive Search Region Adjustment (SRA) strategy is proposed, optimizing the field of view in response to the tracking scenario.
Furthermore, to mitigate the impact of varying target feature proportions introduced by dynamic search factors, 
the Attention-to-Mask (ATM) module—originally developed for object segmentation—is incorporated. This module enhances fine-grained feature representations and integrates hierarchical information, effectively emphasizing foreground targets.
We name our proposed algorithm FocusTrack, which integrates the SRA strategy and the ATM module to keep the target effectively focused within the field of view, thus effectively tackling the challenges of Anti-UAV tracking.

Specifically, the search region adjust (SRA) strategy consists of two key components. First, the SRA module introduces a CLS\_Token that predicts the target presence probability in the field of view. Based on the classical Transformer-based local tracker OSTrack, which learns feature similarity through Attention mechanisms with template and search frames as dual inputs, the CLS\_Token interacts with both during feature extraction. By training on both positive and negative pairs, the CLS\_Token accurately predicts target presence, guiding search factor adjustments. If the probability indicates target absence, the search factor expands, and once the target is reacquired, it returns to the base scale for precise tracking. This dynamic adjustment ensures efficiency in simple scenarios and global search capability in challenging ones.

The Attention-to-Mask (ATM) module, inspired by SegViT\cite{segvit}, is composed of stacked ATM blocks. Each block employs a cross-attention mechanism, where a learnable class token queries search features from backbone layers. The resulting similarity maps, activated by Sigmoid, produce class-specific masks to enhance target discrimination. By aggregating masks from multiple backbone layers, the ATM module constructs a hierarchical feature representation, further refining feature quality.

Experimental results show that our approach achieves state-of-the-art performance with a significant advantage over local-based trackers on two challenging Anti-UAV benchmarks: AntiUAV\cite{antiuav310} and AntiUAV410\cite{antiuav410}, achieving 67.7\% AUC and 62.8\% AUC, respectively. 
In terms of efficiency, FocusTrack shows a clear advantage over global-based trackers with approximately 30G MACs, achieving 143 fps with FocusTrack (SRA) and 44 fps with the full version FocusTrack, both enabling real-time tracking. 
Extensive ablation experiments validate the effectiveness of our model.

The contributions can be summarised as follows:
\begin{itemize}
\item FocusTrack is introduced as a novel framework that combines the Search Region Adjustment (SRA) strategy with the Attention-to-Mask (ATM) module, effectively bridging the gap between local and global trackers for Anti-UAV tracking.

\item An adaptive Search Region Adjustment (SRA) strategy is proposed, dynamically optimizing the field of view based on tracking confidence. The carefully designed CLS\_Token outputs the target presence probability, guiding the adjustment of the search factor during inference.

\item Extensive experimental results confirm the state-of-the-art performance of FocusTrack and its balance between accuracy and efficiency.

\end{itemize}

The remainder of the article is arranged as follows. Section 2 reviews related tracking algorithms and search region adjustment strategies. Section 3 presents our FocusTrack framework, covering the Transformer-Based structure, Search Region Adjustment strategy, Attention-to-Mask module, and training objectives. Section 4 details experimental results, including implementation details, benchmark comparisons, efficiency analysis, ablation studies, and qualitative evaluation. Section 5 concludes with our contributions and future directions.

\vspace{-4mm}
\section{Related Work}
\label{related_work}
\subsection{Single Object Tracking.}
In recent years, single object tracking (SOT) algorithms have advanced rapidly and can be broadly classified into Siamese-based\cite{siamfc, siamfcpp, siamrpn++, huang2023searching} and Transformer-based approaches\cite{ostrack, simtrack, mixformer, artrack, swintrack},. Among them, Transformer-based methods have emerged as the dominant paradigm due to their superior performance. OSTrack\cite{ostrack} and SimTrack\cite{simtrack} pioneered the use of the Vision Transformer (ViT) architecture to simultaneously extract and fuse features from both the template and search region, laying the foundation for subsequent innovations. Moreover, pre-training techniques have been extensively validated as a crucial strategy for enhancing tracking performance\cite{simtrack, vit_survey}. DropMAE\cite{droptrack}, an autoencoder pre-training method specifically designed for matching-based tasks, has demonstrated superior fine-tuning performance on SOT. Therefore, our approach adopts OSTrack\cite{ostrack} as the base network (excluding the Early Candidate Elimination module) and integrates DropMAE\cite{droptrack} pre-trained parameters into the ViT structure to further improve performance.

From the perspective of tracking strategies, SOT methods can be categorized into local-based tracking\cite{siamfc, ostrack, mixformer, artrack} and global-based tracking\cite{globaltrack, siamrcnn, antiuav410}. 
Local-based tracking assumes that the target's position and scale vary smoothly across consecutive frames, thus limiting the search to a local region to reduce computational overhead. However, this approach is prone to tracking failure when the target undergoes abrupt motion or occlusion. 
In contrast, global-based tracking searches for the target across the entire image, making it more robust against target loss but significantly increasing computational cost, making real-time tracking infeasible. 
Our method dynamically balances local and global tracking by adaptively adjusting the local tracker's search field of view, simultaneously optimizing tracking stability and computational efficiency.

\vspace{-3mm}
\subsection{Anti-UAV Tracking.}
Anti-UAV vision-based tracking is crucial in counter-UAV technology, with infrared modality receiving extensive attention due to its all-weather adaptability~\cite{survey}. 
Existing datasets, such as AntiUAV~\cite{antiuav310} and AntiUAV410~\cite{antiuav410}, are specifically designed for real-world UAV tracking scenarios, where AntiUAV~\cite{antiuav310} provides infrared-visible dual-modality data, while AntiUAV410~\cite{antiuav410} focuses solely on infrared modality.  

Anti-UAV tracking faces significant challenges, including Severe Drift (\emph{e.g.}, full occlusions, out-of-view) and Drastic Motion (\emph{e.g.}, rapid movements, viewpoint changes). Consequently, most existing methods adopt global-based tracking strategies. SiamSTA~\cite{siamsta} and GASiam~\cite{gasiam} are built upon SiamR-CNN\cite{siamrcnn}, leveraging a three-stage global re-detection mechanism and an adaptive local-global switching strategy to handle target loss. Specifically, SiamSTA~\cite{siamsta} employs spatio-temporal attention to refine candidate regions and enhance small target perception, while GASiam~\cite{gasiam} utilizes graph attention to improve feature embedding in local tracking networks, mitigating background interference. In contrast, SiamDT~\cite{antiuav410} also follows a global tracking paradigm but introduces a dual-semantic feature extraction mechanism along with a background distractor suppression strategy to enhance UAV target discrimination in dynamic backgrounds. These methods have demonstrated state-of-the-art performance on Anti-UAV datasets, providing effective solutions for UAV tracking in complex scenarios.  

However, global tracking methods impose substantial computational costs, typically running below 10fps, making real-time tracking infeasible.
To address this limitation, we explore solutions based on local trackers. However, local trackers have shown subpar performance in the Anti-UAV task, and no existing method has explored local-based Anti-UAV tracking. 
A systematic analysis of their limitations reveals key weaknesses, guiding the development of targeted enhancements. This work pioneers the first local-based Anti-UAV tracking framework, bridging the gap between efficiency and performance.

\vspace{-3mm}
\subsection{Search Region Adjustment Strategy in Anti-UAV Tracking.}

In local search mechanisms, the search region is obtained via a crop-and-resize approach, where an area around the target’s previous position is sampled. The size of this region is determined by multiplying the target size by a search factor, followed by resizing it to a predefined dimension. Given the previous position $b(b^{cx},~b^{cy},~b^{w},~b^{h})$ and the search factor $f$, the crop size $(W, H)$ is computed as:  
$W_i = H_i = \lceil f_i \times \sqrt{b_i^w \times b_i^h} \rceil, \quad i \in \{\text{template, search}\}$, where $\lceil \cdot \rceil$ denotes the ceiling function. 
The search factor $f$ controls the amount of background context in the visual field, the larger the value, the larger the field of view and vice versa. 
Typically, $f_{\text{template}} = 2$, whereas $f_{\text{search}}$ depends on the input resolution: $f_{\text{search}} = 4$ for an input size of $256$, and $f_{\text{search}} = 5$ for an input size of $384$~\cite{ostrack, mixformer, droptrack}.

In general single-object tracking (SOT) tasks, a fixed search factor of 4 or 5 is sufficient to keep the target within view, as objects in SOT are usually large and exhibit relatively smooth motion. As a result, most existing tracking methods rely on a fixed search factor without further refinement, and research on search region adjustment strategies remains scarce.  

However, Anti-UAV tracking presents unique challenges, and our experiments reveal that the choice of search region significantly affects tracking performance. This makes the introduction of a Search Region Adjustment Strategy both necessary and well-founded for Anti-UAV tracking. Extensive experiments validate the effectiveness of our approach.

\section{Methods}
\label{method}

\begin{figure*}[ht]
\centering
\includegraphics[scale=0.55]{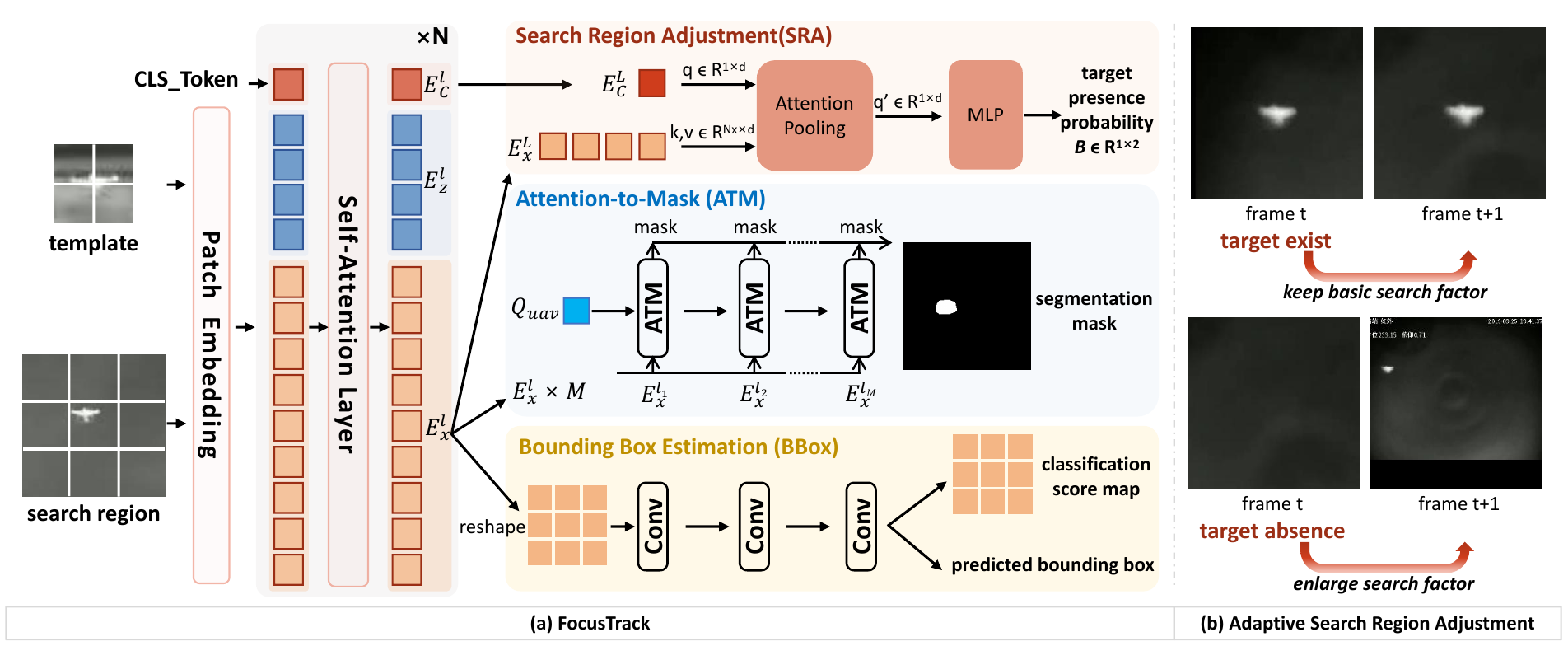} 
\caption{
(a) Overall structure of FocusTrack. (b) Detailed explanation of the Adaptive Search Region Adjustment Strategy. FocusTrack takes the template, search region, and a learnable \texttt{CLS\_Token} as inputs, processing them through Patch Embedding and multiple self-attention layers for feature extraction. The extracted features then pass sequentially through the Search Region Adjustment, Attention-to-Mask, and Bounding Box Estimation modules, ultimately producing the target presence probability, segmentation mask, classification score map, and predicted bounding box.}
\label{fig:framework} 
\end{figure*}

For Anti-UAV tracking, FocusTrack (illustrated in Fig. \ref{fig:framework} (a)) is introduced as a solution that maintains the target within the field of view through adaptive search region adjustments. This approach bridges the gap between local and global tracking, achieving both high accuracy and speed.
Specifically, the proposed adaptive Search Region Adjustment (SRA) strategy adjusts the search factor by evaluating the probability of target presence, and the proposed Attention-to Mask (ATM) module effectively addresses the feature degradation caused by the increase of search factor by introducing hierarchical structure and refined features.

In the following, we first revisit the basic structure in Transformer-based local tracking algorithm, then detail the SRA strategy and ATM module, and finally specify the loss function for model training.

\subsection{Transformer-Based Local Tracking Structure.}
\label{method:revisiting}

\subsubsection{Unified Feature Extraction}

Transformer-based local tracking algorithms~\cite{mixformer, ostrack, simtrack} have demonstrated outstanding performance in single object tracking by effectively capturing the similarity between the template frame $\boldsymbol{Z} \in \mathbb{R}^{h_{z} \times w_{z} \times 3}$ and search frame $\boldsymbol{X} \in \mathbb{R}^{h_{x} \times w_{x} \times 3}$ through attention mechanisms.
Under the local tracking paradigm, the model processes cropped regions that are resized to fixed dimensions, yielding a template region $\boldsymbol{z} \in \mathbb{R}^{H_{z} \times W_{z} \times 3}$ and a search region $\boldsymbol{x} \in \mathbb{R}^{H_{x} \times W_{x} \times 3}$ as model input. 

These regions are then encoded through a patch embedding process, as illustrated in Fig.\ref{fig:framework} (a) (Patch Embedding module), which follows the same strategy as used in Vision Transformers (ViT)\cite{vit}.
Specifically, the process is implemented using a convolutional layer with kernel size $P \times P$, stride $P$, and $C$ output channels, which simultaneously partitions the input into non-overlapping patches and maps them into a $C$-dimensional space. As a result, the template and search regions yield $N_{z} = H_{z}W_{z} / P^{2}$ and $N_{x} = H_{x}W_{x} / P^{2}$ patches, respectively, producing the embedded features $\mathbf{E}_{z} \in \mathbb{R}^{N_{z} \times C}$ and $\mathbf{E}_{x} \in \mathbb{R}^{N_{x} \times C}$ as transformer input tokens. Notably, the patch embedding layers for the template and search regions share the same parameters, ensuring a consistent feature representation across both inputs.

To incorporate spatial information, learnable positional embeddings are added to $\mathbf{E}_{z}$ and $\mathbf{E}_{x}$, forming the initial token representations. 
Subsequently, the template and search tokens are concatenated into a unified sequence of length $N_{z} + N_{x}$, serving as identical query ($\boldsymbol{q}$), key ($\boldsymbol{k}$), and value ($\boldsymbol{v}$) inputs to the feature extraction module.
This module consists of $L$ stacked self-attention layers, each comprising a multi-head attention (MHA) module followed by a feed-forward network (FFN), both employing residual connections. The operations within the $l$-th layer are formally defined as:
\begin{equation}
\begin{aligned}
    \boldsymbol{q}  = \boldsymbol{k} = \boldsymbol{v} & = [\mathbf{E}^{l-1}_{z}; \mathbf{E}^{l-1}_{x}], \\ 
    [\mathbf{E'}^{l-1}_{z}; \mathbf{E'}^{l-1}_{x}] & = [\mathbf{E}^{l-1}_{z}; \mathbf{E}^{l-1}_{x}] + \text{MHA}(\boldsymbol{q}, \boldsymbol{k}, \boldsymbol{v}), \\
    [\mathbf{E}^{l}_{z}; \mathbf{E}^{l}_{x}] & = [\mathbf{E'}^{l-1}_{z}; \mathbf{E'}^{l-1}_{x}] + \text{FFN}([\mathbf{E'}^{l-1}_{z}; \mathbf{E'}^{l-1}_{x}]),
\end{aligned}
\end{equation}
where $[;]$ denotes concatenation, and $\mathbf{E}^{l-1}_{z}$, $\mathbf{E}^{l-1}_{x}$ represent the input tokens for the $l$-th layer. This unified processing enables comprehensive feature extraction through both self-attention within each region and cross-attention between the template and search regions, facilitating robust object tracking.

\subsubsection{Bounding Box Estimation}

After passing through all encoder layers, the search tokens are extracted from the combined sequence and rearranged into their original spatial configuration, forming a 2D feature map $\boldsymbol{E_{x}} \in \mathbb{R}^{ C \times \frac{H_x}{P} \times \frac{W_x}{P}}$, which serves as the input for Bounding Box estimation(short for BBox in Fig.\ref{fig:framework} (a) ). 

Specifically, this feature map is processed by a stack of convolutional layers designed for target localization and bounding box regression~\cite{ostrack, mcitrack}. The module outputs three key components: the classification score map $\boldsymbol{P} \in[0,1]^{\frac{H_x}{P} \times \frac{W_x}{P}}$, the local offset $\boldsymbol{O} \in[0,1)^{2 \times \frac{H_x}{P} \times \frac{W_x}{P}}$, and the normalized bounding box size (i.e., width and height) $\boldsymbol{S} \in[0,1]^{2 \times \frac{H_x}{P} \times \frac{W_x}{P}}$. 
The target position $(x_d, y_d)$ is determined by the location with the highest classification score in $\boldsymbol{P}$. The final predicted bounding box is then computed using the corresponding local offset $\boldsymbol{O}_{(x_d, y_d)}$ and bounding box size $\boldsymbol{S}_{(x_d, y_d)}$.

\begin{figure}[t]
\centering
\includegraphics[width=0.45\textwidth]{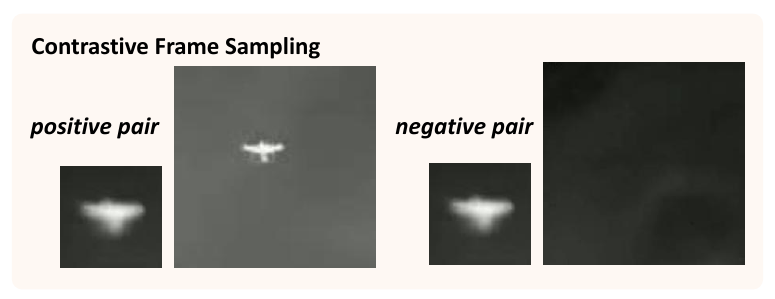}
\vspace{-2mm}
\caption{
 Detailed explanation of the Contrastive Frame Sampling Strategy.
}
\label{fig:sampling}
\end{figure}

\vspace{-2mm}
\subsection{Adaptive Search Region Adjustment Strategy.}
\label{method:rsa}
Building upon the Transformer-based local tracking framework, an adaptive Search Region Adjustment (SRA) strategy is introduced to dynamically adjusts the search region to maintain target visibility, effectively handling scenarios where the target moves out of the field of view.
As illustrated in Fig. \ref{fig:framework}(b), the SRA module first estimates the probability of the target remaining within the visible range. 
Based on this estimation, the search factor \( f_x \) is adaptively adjusted during inference, enabling a dynamic tracking view.

\subsubsection{Search Region Adjustment Module}
To enable robust target presence estimation, the architecture incorporates a learnable {CLS\_Token} $\mathbf{E}_{c} \in \mathbb{R}^{1 \times C}$ at the input layer, inspired by ViT\cite{vit} and BERT\cite{bert} frameworks. This token, enriched with positional embedding and concatenated with template and search tokens ($\mathbf{E}_{z} \in \mathbb{R}^{N_{z} \times C}$ and $\mathbf{E}_{x} \in \mathbb{R}^{N_{x} \times C}$), feeds into the backbone network. Through extensive feature interactions, the output {CLS\_Token} $\mathbf{E}^{L}_{c} \in \mathbb{R}^{1 \times C}$ evolves to capture high-dimensional representations that characterize the template-search relationship.
Building upon these enriched features, the SRA module employs a two-stage processing pipeline. 
First, it performs feature enhancement through Attention Pooling, which is essentially a single-layer cross-attention mechanism. In this process, the {CLS\_Token} $\mathbf{E}^{L}{c} \in \mathbb{R}^{1 \times C}$ serves as the \emph{query}, while the output search tokens $\mathbf{E}^{L}{x} \in \mathbb{R}^{N_{x} \times C}$ serve as both the \emph{key} and \emph{value}. Then, a two-layer MLP adjusts the channel dimensions, ultimately producing a binary classification vector $\mathbf{B} \in \mathbb{R}^{1 \times 2}$, where $\mathbf{B}[0]$ (denoted as \textbf{logits}) represents the probability of target presence.

\begin{figure}[t]
    \centering
    \includegraphics[width=0.46\textwidth]{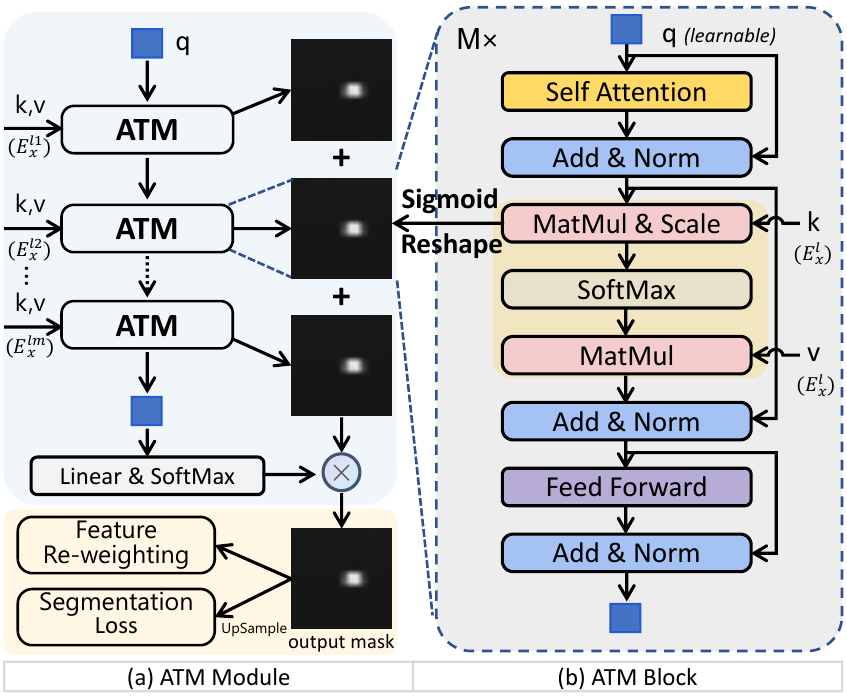}
    \vspace{-2mm}
    \caption{(a) Overall structure of ATM Module. (b) Detailed structure of a single ATM block. The ATM module consists of $M$ stacked ATM blocks, taking a learnable category query and multi-layer search tokens as input to generate the target segmentation mask.
}
\vspace{-2mm}
    \label{fig:atm}
\end{figure}

\subsubsection{Contrastive Frame Sampling}
To enhance the training of the SRA module, a contrastive sampling strategy (as shown in Fig.~\ref{fig:sampling}) extends beyond the conventional method of selecting template and search frames from the same sequence (positive pairs, target presence probability = 1). Negative pairs are additionally introduced by selecting frames from different sequences (negative pairs, target presence probability = 0). This contrastive sampling strategy enables effective supervision through the CrossEntropy loss, denoted as \textbf{\(L_{\text{logits}}\)}.

\subsubsection{Adaptive Search Region Adjustment}
During inference, the presence logits dynamically guide search factor \( f_{x} \) adjustment.
Specifically, if the logits exceed a threshold \( T_{\text{logits}} \), signifying that the target remains within the current view, the search factor retains its base value \( f^{\text{base}}_{x} \) for the next frame. Conversely, if the logits fall below the threshold, suggesting target absence, the search factor is incrementally increased to expand the view field until the target is reacquired. Upon successful reacquisition, the search factor reverts to \( f^{\text{base}}_{x} \) to maintain tracking stability.  
To prevent excessive expansion, 
a stepwise increase in the search factor is employed, constrained by an empirically determined maximum value \( f^{\text{max}}_{x} \), beyond which no further expansion is applied.

Additionally, the Bounding Box Estimation Module (Section \ref{method:revisiting}) inherently adapts to target motion by predicting appropriately scaled bounding boxes, reducing the frequency of search region adjustments.   
To further enhance both robustness and computational efficiency, the peak score  \( P_{max} \) from the classification map \( P \) serves as an additional criterion. A high peak score indicates strong classification confidence, while a low score suggests uncertainty. The search factor is updated only when both the presence probability and peak score fall below their respective thresholds, \( T_{\text{logits}} \) and \( T_{\text{score}} \). The adaptive adjustment strategy is detailed in the following pseudo-code:
\begin{algorithm}
\caption{Adaptive Search Region Adjustment}
\begin{algorithmic}[1] 
\Require logits, $P_{max}$, step, $f_{max}$, $f_{base}$
\Ensure search factor $f$
\If{use\_region\_adjust enabled}
    \If{logits $< T_{\text{logits}}$ \textbf{and} $P_{max} < T_{\text{score}}$}
        \State $f \gets \min(f + \text{step}, f_{max})$
    \Else
        \State $f \gets f_{base}$
    \EndIf
\EndIf
\end{algorithmic}
\end{algorithm}

\vspace{-4mm}
\subsection{Attention-to-Mask Module for Feature Refinement.}
\label{method:atm}

To mitigate feature degradation caused by an increased search factor, the Attention-to-Mask (ATM) module is introduced, comprising multiple stacked ATM blocks.

\subsubsection{Attention-to-Mask Block}

As shown on the right side of Fig. \ref{fig:atm} , the ATM block functions as a target mask extractor based on Cross-Attention mechanism. 
It utilizes a learnable class token $\mathbf{Q}_{uav} \in \mathbb{R}^{1 \times C}$ as the query ($\mathbf{q}$), 
while using search tokens from each backbone layer $\mathbf{E}^{l}_{x} \in \mathbb{R}^{N_{x} \times C}$ as keys ($\mathbf{k}$) and values ($\mathbf{v}$). During Multi-head Attention computation, the dot-product between query and keys generates a similarity map $\mathbf{S}$:
\begin{equation} 
\begin{split}
       S(q, k)& = \frac{qk^T}{\sqrt{d_{k}}} \in \mathbb{R}^{1 \times N_{x}}.
\end{split}
\end{equation}
where higher values indicate stronger correlations with UAV targets. 
The resulting similarity map, after Sigmoid activation, can be further reshaped to $1 \times {H_x}/{P} \times {W_x}/{P} $, which effectively serves as a target mask.

\subsubsection{Attention-to-Mask Module}

Vision Transformer (ViT) layers inherently capture hierarchical semantic representations, with each layer contributing distinct and increasingly high-level feature information~\cite{vit}.
The ATM module capitalizes on this property by integrating multiple stacked ATM blocks, enabling multi-layer feature fusion and enhancing mask robustness.

As shown in Fig.~\ref{fig:atm}, the module sequentially processes search tokens $\mathbf{E}^{l}_{x} \in \mathbb{R}^{N_{x} \times C}$ from multiple backbone layers ($l \in [1,2,...,L]$), where each block refines its input and passes enriched features forward, progressively enhancing representation quality. This hierarchical process produces two key outputs: a category token encoding UAV class information and multiple UAV segmentation masks. The category token undergoes a Linear-Softmax transformation to generate class scores. Inspired by SegVit~\cite{segvit}, an additional background class is introduced to handle target-absent scenarios. 
The generated masks, maintaining spatial alignment and semantic consistency, are fused via element-wise summation.

Unlike conventional segmentation methods that separately supervise category tokens and masks, our single-class UAV segmentation task allows for a streamlined supervision strategy. 
The fused mask, weighted by class scores, serves a dual purpose: (1) acting as an attention weight to refine the classification score map from the Bounding Box Estimation Module (Section \ref{method:revisiting}), suppressing background noise while emphasizing foreground features; 
(2) functioning as a pixel-wise segmentation map, which is upsampled to align with the search region and optimized using Focal Loss, denoted as \textbf{$L_{\text{mask}}$}.

\vspace{-2mm}
\subsection{Training Objective.}
The overall training objective integrates multiple loss functions into a unified framework. 
For the Bounding Box Estimation Module, Focal Loss~\cite{focalloss} is employed for classification, while L1 Loss and GIoU Loss~\cite{giou} handle regression. 
The SRA module incorporates Cross Entropy Loss to supervise the target presence probability, while the ATM module applies Focal Loss for target mask supervision. The final total loss function is formulated as:
\begin{equation}
    L = \lambda_{\text{giou}} L_{\text{giou}} + \lambda_{\text{L1}} L_{\text{L1}} + L_{\text{focal}} + L_{\text{logits}} + L_{\text{mask}},
\end{equation}
\vspace{-1mm}
where $\lambda_{\text{giou}} = 2$, $\lambda_{\text{L1}} = 5$, following settings in OSTrack \cite{ostrack}.

\section{Results}
\label{results}

% In this section, we first present the implementation details of FocusTrack, followed by state-of-the-art and efficiency comparisons, demonstrating that our tracker effectively balances accuracy and computational efficiency. Subsequently, we conduct ablation experiments, attribute-based analysis, and qualitative analysis to comprehensively validate the effectiveness of our approach.

\subsection{Implementation Details.}

Our tracker is implemented in Python 3.9 and PyTorch 2.0.0, with experiments conducted on dual 24GB NVIDIA RTX 3090 GPUs.

\subsubsection{Model}
FocusTrack comprises a ViT-B backbone, SRA module, ATM module, and a center-based bounding box estimation module. 
The backbone leverages DropMAE pretraining\cite{droptrack}, a specialized masked autoencoder optimized for matching-based tasks, to enhance temporal correspondence learning. 
To ensure compatibility with this pretraining approach, learnable frame identity embeddings are incorporated, distinguishing template and search representations.
The Attention Pooling in SRA module employs a single-layer cross-attention mechanism with a hidden dimension of 768 and 8 attention heads.
The ATM module consists of 3 stacked ATM blocks, each receiving the outputs from the 6th, 8th, and 12th layers of the backbone as key and value inputs. Each ATM block encompasses 3 stacked cross-attention layers with a hidden dimension of 384 and 8 attention heads--a configuration aligned with SegVit's architecture. 
All other module parameters except backbone are randomly initialized.

\subsubsection{Dataset and Preprocessing}
The model is trained on the AntiUAV410 train set~\cite{antiuav410}. 
Based on empirical search factor analysis from OSTrack, search factors for the template and search regions are set to 2 and 6, resizing them to 128×128 and 256×256, respectively. 
For segmentation supervision, rectangular pseudo ground-truth masks are generated from bounding box annotations, serving as a reasonable approximation given the small size of UAVs and their predominantly axis-aligned flight postures. Standard augmentations such as horizontal flipping and brightness jittering enhance training robustness.

\subsubsection{Training}
Two training paradigms are explored to balance performance and training efficiency:

\noindent 1) Single-Phase Training: This approach processes positive and negative samples simultaneously at a 7:3 ratio, training all modules (SRA, ATM, and BBox) concurrently. Each iteration processes 60,000 positive samples, so the total number of samples per iteration is \( 60,000 / 0.7 = 85,715 \). The model trains for 30 epochs with learning rates of $4 \times 10^{-5}$ for the backbone and $4 \times 10^{-4}$ for auxiliary modules.
   
\noindent 2) Two-Phase Training: 
Recognizing that only the SRA module requires negative samples, an efficient two-phase approach is implemented: 
\\
    \underline{\textbf{\textit{Phase 1:}}} Train a preliminary model without SRA module using 60,000 positive samples per iteration, optimizing ATM and BBox parameters. Training spans 20 epochs with identical learning rates as single-phase training.\\
    \underline{\textbf{\textit{Phase 2:}}} Introduce negative samples for SRA module optimization while freezing all other parameters. Sample distribution matches single-phase training, with the learning rate of ATM module set to $4 \times 10^{-5}$.

\input{tables/comp_on_antiuav310}

\subsubsection{Inference}

During the inference phase, the base search factor \( f^{\text{base}}_{x} \) is set to 6, the step for increasing the search factor \( f^{\text{step}}_{x} \) is 1, and the maximum search factor \( f^{\text{max}}_{x} \) value is 8. The threshold for logits is  \( T_{logits}= 0.8 \) , and the threshold for score is \( T_{score}= 0.5 \). A Hanning window penalty is adopted to utilize positional prior in tracking, following the common practice~\cite{ostrack, transt}. At the same time, when the search region needs adjustment, the Hanning window penalty is disabled.

\subsubsection{Evaluation Metrics}

Tracking performance is evaluated using four key metrics: success rate, precision, normalized precision, and state accuracy~\cite{antiuav310}. The success rate uses the IoU between predicted and ground truth boxes, with the Area Under Curve (AUC) of the success plot serving as the overall accuracy metric. The precision measures center location error, with precision score (P) at 20 pixels as the location accuracy measure. Normalized precision (Pnorm) accounts for target size variations by normalizing the center error with target dimensions. 
Additionally, state accuracy (SA), proposed in ~\cite{antiuav310}, evaluates both localization accuracy and visibility prediction, providing a more comprehensive assessment of tracking performance in Anti-UAV tracking scenarios.

\vspace{-3mm}
\subsection{State-of-the-art Comparison.}
The proposed method is evaluated on two challenging Anti-UAV benchmarks: AntiUAV~\cite{antiuav310} and AntiUAV410~\cite{antiuav410}. 
To ensure a fair comparison, existing results from~\cite{antiuav410, promptvt} are collected, and four state-of-the-art transformer-based local trackers—OSTrack~\cite{ostrack}, ROMTrack~\cite{romtrack}, ZoomTrack~\cite{zoomtrack}, and DropTrack~\cite{droptrack}—are retrained on the AntiUAV410 training set using their default configurations and pre-trained weights, ensuring consistent experimental conditions.

\input{tables/comp_on_antiuav410}

\subsubsection{Evaluation on AntiUAV}

The AntiUAV dataset consists of over 318 video pairs with more than 580K manually annotated bounding boxes, covering both visible and thermal infrared test videos. Our method is evaluated on the thermal infrared modality.

Table~\ref{tab:antiuav310} presents the performance comparison between our tracker and state-of-the-art methods. All reported results are obtained from models trained on the AntiUAV410 training set\cite{antiuav410}. The evaluation demonstrates that our tracker consistently outperforms existing methods across all four metrics, achieving absolute superiority in tracking performance.

\subsubsection{Evaluation on AntiUAV410}

The AntiUAV410 dataset comprises 410 thermal infrared sequences with 438K+ annotated frames, partitioned into train (200 sequences), validation (90 sequences), and test (120 sequences) sets. With an average sequence length of 1,069 frames, the dataset emphasizes challenging scenarios: over 50\% of targets are smaller than 50 pixels, including instances below 10 pixels. 

A comprehensive evaluation of 18 transformer-based local trackers is conducted on the AntiUAV410 test set, as shown in Table~\ref{tab:antiuav410}.
These trackers are divided into two groups: the upper section reports results obtained by evaluating on their original models, while the lower section presents results from models retrained on AntiUAV410 test set using default configurations and pre-trained weights. 

\begin{figure*}[t]
\begin{center}
\subfigure{\includegraphics[width=0.30\textwidth]{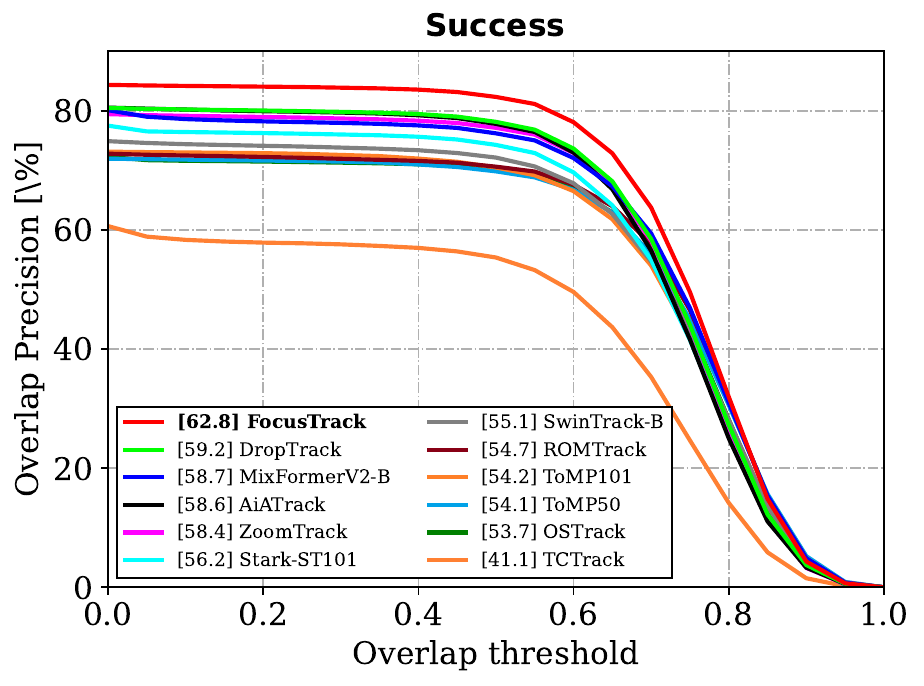}}\hfill
\subfigure{\includegraphics[width=0.30\textwidth]{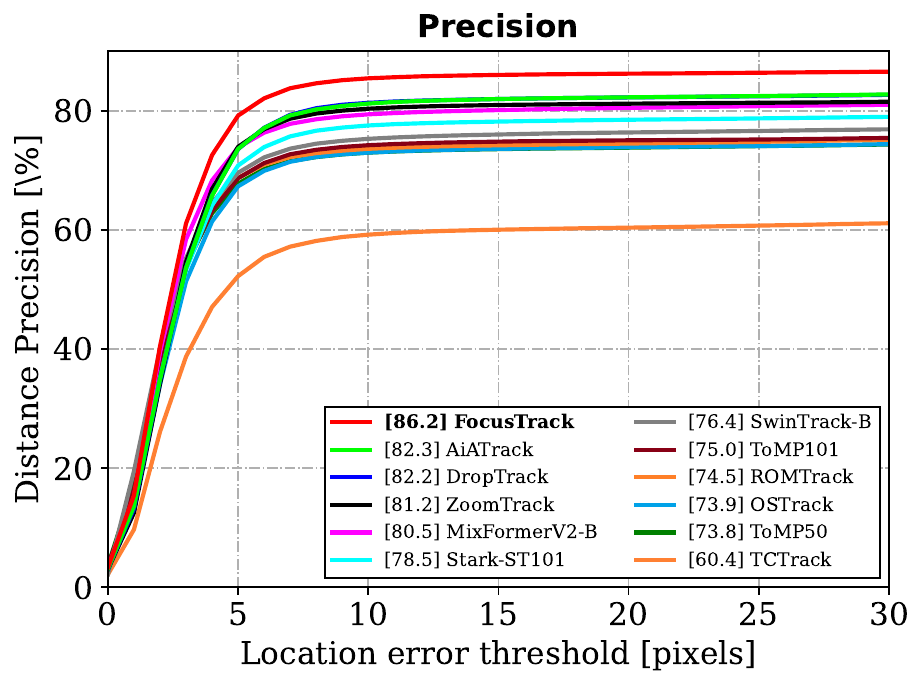}}\hfill
\subfigure{\includegraphics[width=0.30\textwidth]{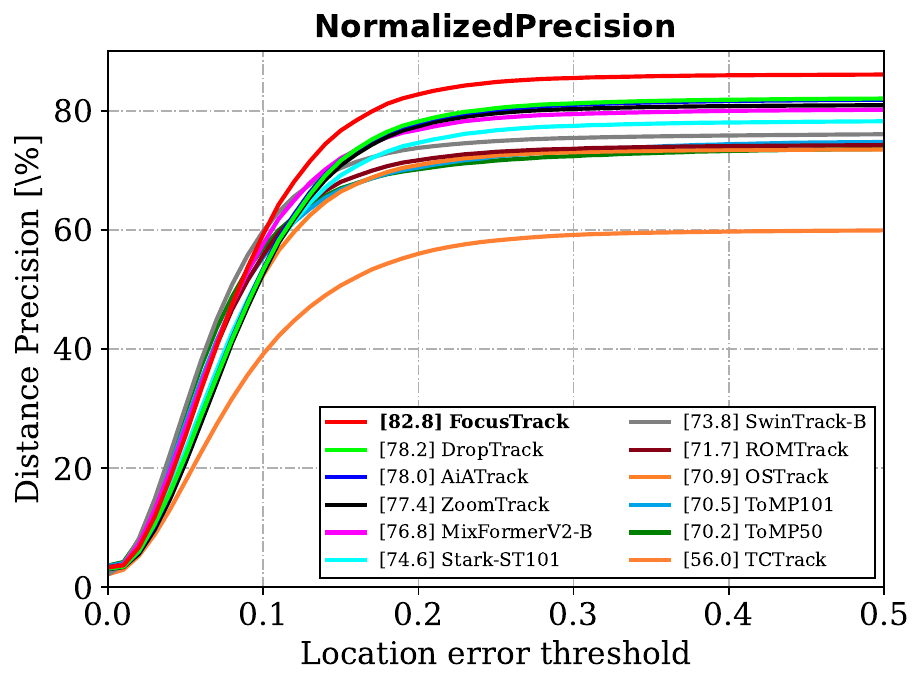}}
\end{center}
\vspace{-5mm}
\caption{Success, Precision and Normalized Precision plots on AntiUAV410 test set.}
\vspace{-4mm}
\label{fig:antiuav410_plot}
\end{figure*}

Notably, our proposed FocusTrack consistently outperforms all competitors across all metrics.
This superiority is further illustrated in Fig.~\ref{fig:antiuav410_plot}, which visualizes the success rate, precision, and normalized precision plots for the retrained trackers.

\input{tables/fps}

\vspace{-3mm}
\subsection{Efficiency Comparison.}

Table~\ref{tab:efficiency} compares FocusTrack with state-of-the-art transformer-based local and global trackers in terms of speed (fps), computational complexity (MACs), and tracking performance (AUC, P, SA). All speed evaluations are conducted on a single NVIDIA RTX 3090 GPU.

The variant of FocusTrack excluding the ATM module, referred to as FocusTrack (SRA), demonstrates a strong balance between efficiency and performance, achieving 62.3\% AUC while maintaining a high speed of 143 fps with minimal computational overhead (29.1G MACs). 
The full version of FocusTrack, incorporating the ATM module, further enhances performance to 62.8\% AUC at the cost of reduced speed (44 fps). Despite this trade-off, it remains real-time capable while introducing additional mask prediction functionality.

Compared to the transformer-based global tracker SiamDT, which achieves the highest performance (66.8\% AUC), FocusTrack demonstrates superior efficiency, operating over 5$\times$ faster and requiring approximately 1/8 of the computational cost, while maintaining competitive performance (4.0\% lower in AUC). This efficiency-performance trade-off reinforces our design choice of adopting a local tracking paradigm, which effectively balances accuracy and computational efficiency.

\vspace{-2mm}
\subsection{Ablation Experiment.}

To assess the impact of each component within FocusTrack, comprehensive ablation experiments are conducted on the AntiUAV410\cite{antiuav410} test set.

\subsubsection{Effectiveness of Each Component}
Table~\ref{tab:ablation} presents a progressive integration of various components into the baseline model, revealing their individual contributions. 
Starting from our baseline model OSTrack (53.7\% AUC), introducing a search factor of 6 (sf=6) brings a significant improvement of 6.1\% in AUC, and the integration of DropMAE pre-trained weights further enhances performance by 0.4\%.

\input{tables/ablation}

The ablation results demonstrate the effectiveness of our key modules. Specifically, incorporating the SRA module improves performance to 62.3\% AUC (+8.6\%), highlighting its crucial role in adapting the search region based on target state awareness. This substantial gain validates our design choice of dynamic search region adjustment. The ATM module alone achieves 60.6\% AUC (+6.9\%), demonstrating its effectiveness in refining feature representations through hierarchical attention-guided mask learning.
When combining both SRA and ATM modules, our model achieves 62.6\% AUC, representing a significant improvement of 8.9\% over the baseline. This synergistic effect indicates that the two modules complement each other effectively, with SRA providing precise target localization and ATM enhancing feature discrimination. Finally, the proposed two-phase training strategy further boosts the performance to 62.8\% AUC, validating the effectiveness of our complete FocusTrack framework.

\input{tables/ablation_sra_training}

\input{tables/ablation_sra_inference}

\subsubsection{Discussion on Search Region Adjustment Module}

To thoroughly evaluate our Search Region Adjustment (SRA) module, ablation studies are conducted in both training and inference phases. Note that all experiments are conducted with \underline{single-phase} training strategy for fair comparison.

\noindent\textbf{Training Phase} 
The investigation begins by exploring the optimal architectural design of the SRA module, with a particular focus on feature pooling strategies and the MLP structure. 
Table~\ref{tab:sra_train} presents the results of comparing various pooling methods (row 1-3), while the MLP layers are fixed at two. The comparison involves using the CLS token directly (61.6\% AUC), average pooling (61.7\% AUC), and attention pooling (62.6\% AUC). The results demonstrate that attention pooling significantly outperforms other strategies by effectively capturing the relationship between cls token and search tokens. 

Further analysis is conducted on the impact of MLP depth with attention pooling (row 3-4), where a 2-layer structure (62.6\% AUC) shows clear advantages over a single layer (60.4\% AUC), indicating the necessity of sufficient feature transformation capacity.

\begin{figure*}[b]
\begin{center}
\vspace{-3mm}
\includegraphics[width=0.29\linewidth]{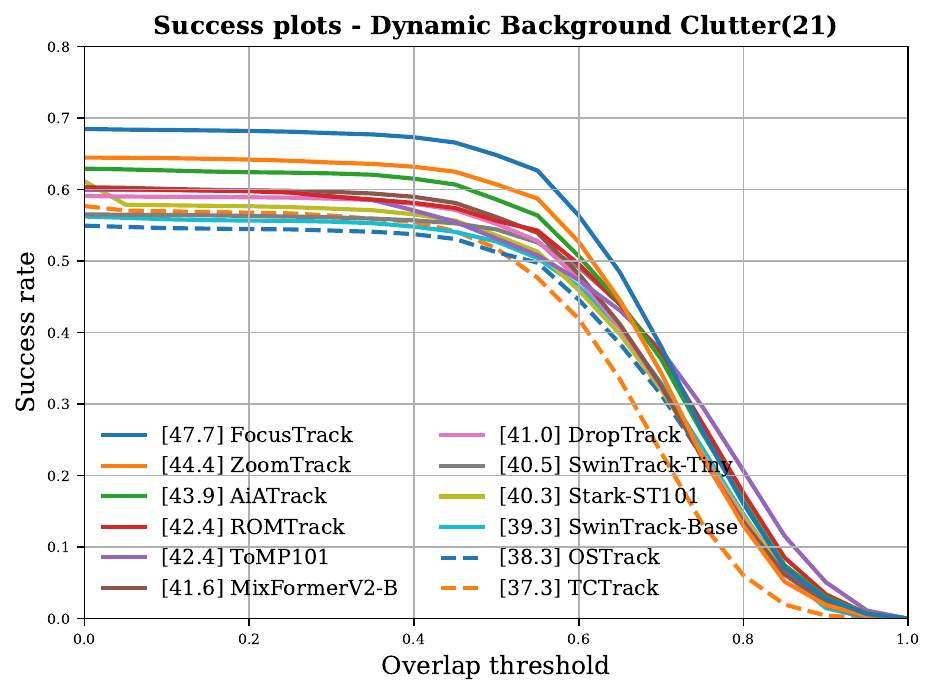}\hfill
\includegraphics[width=0.29\linewidth]{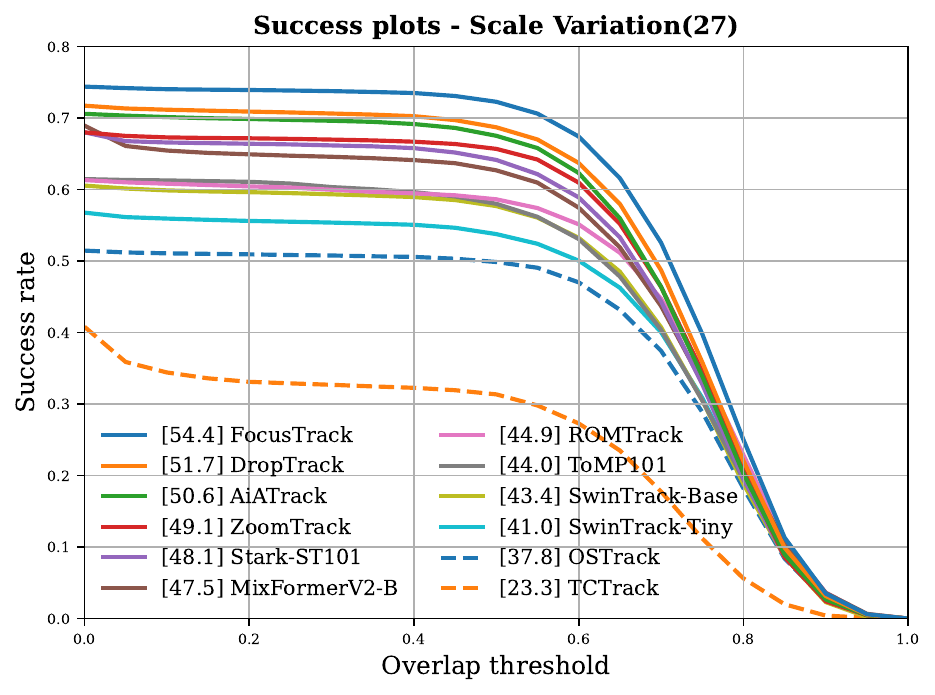}\hfill
\includegraphics[width=0.29\linewidth]{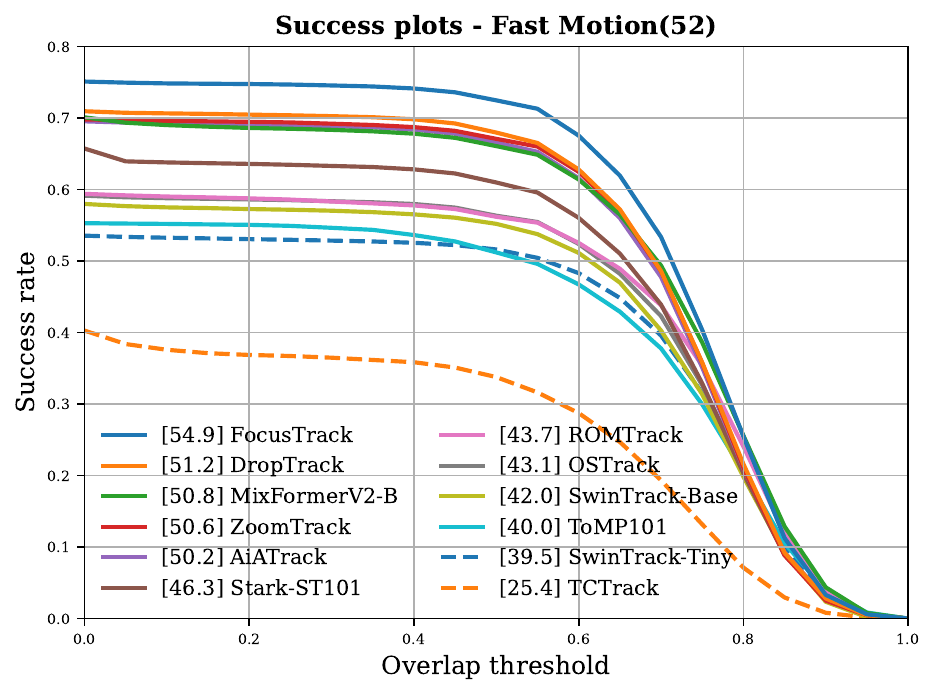}\hfill
\includegraphics[width=0.29\linewidth]{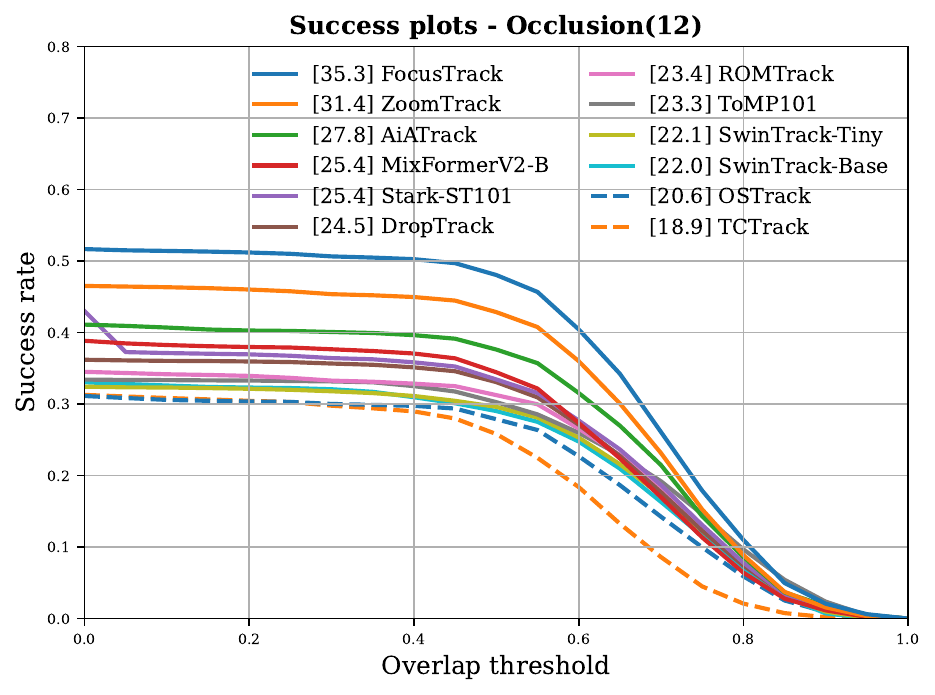}\hfill
\includegraphics[width=0.29\linewidth]{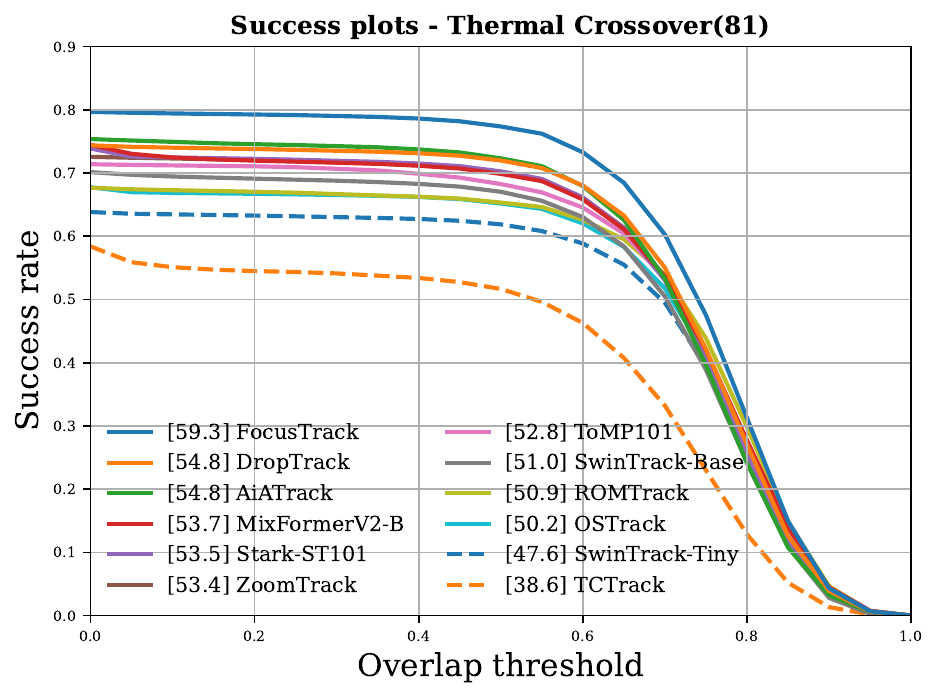}\hfill
\includegraphics[width=0.29\linewidth]{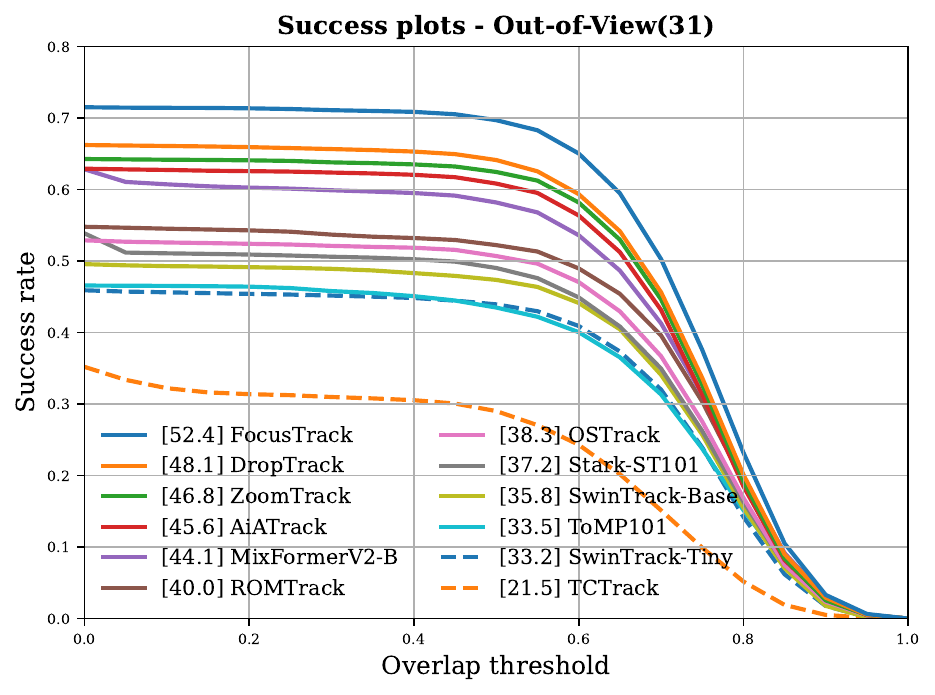}\hfill
\end{center}
\vspace{-4mm}
\caption{Attribute-based success performance analysis on the AntiUAV410 test set.
Success plots are shown for six challenging scenarios: Dynamic Background Clutter (DBC), Scale Variation (SV), Fast Motion (FM), Occlusion (OC), Thermal Crossover (TC), and Out-of-View (OV). Better viewed in color and zoom modes.}
\label{fig:attribute_success}
\end{figure*}

\noindent\textbf{Inference Phase}
In the inference phase, the search region adjustment strategy is influenced by four critical hyperparameters.
As shown in Table~\ref{tab:sra_inference}, a series of controlled experiments were conducted by varying one parameter while fixing others at their optimal values. 
The experimental results reveal optimal values of
\( T_{logits}= 0.8 \), \( T_{score}= 0.5 \), \( f^{\text{max}}_{x}=8 \) and \( f^{\text{step}}_{x}=1 \),
collectively ensuring robust target tracking performance.

\subsubsection{Discussion on Attention-to-Mask Module}

To thoroughly evaluate the attention-to-Mask (ATM) module, ablation studies focus on both the module's structure and its loss function. For consistency, all experiments are conducted using the \underline{single-phase} training strategy.

\noindent\textbf{Module Structure}
The impact of feature hierarchies on the ATM module is examined by varying the number and selection of input layers from the backbone.
As shown in Table~\ref{tab:atm}, progressively increasing the number of stages improves tracking performance, with the three-stage configuration achieving the best results. Among different three-stage combinations, using layers [6, 8, 12] yields the highest AUC of 62.6\%, outperforming other selections. However, extending to four stages degrades performance, suggesting that excessive feature aggregation introduces redundancy.  
\input{tables/ablation_atm}

\noindent\textbf{Loss Function}
An ablation study is also conducted to assess the impact of different loss functions for the segmentation task within the ATM module. 
As shown in the first row of Table~\ref{tab:loss_ablation}, "w/o supervision" refers to the case where no explicit segmentation supervision is applied. Despite the lack of direct supervision, the ATM module still demonstrates reasonable performance. This is primarily due to the module's design, where the generated mask serves a dual purpose: not only as a segmentation output but also as an attention weight that refines the classification score map. Consequently, even in the absence of explicit segmentation loss, the model is able to optimize its parameters through the bounding box estimation pathway.

Rows 2-4 of Table~\ref{tab:loss_ablation} present results for different segmentation loss functions: Dice Loss, Focal Loss, and their combination. Among these, Focal Loss stands out by yielding the best performance, thereby confirming its effectiveness and making it the most suitable choice for segmentation supervision in our final model.
\input{tables/loss_function}

\begin{figure}[h]
    \centering
    \includegraphics[width=0.35\textwidth]{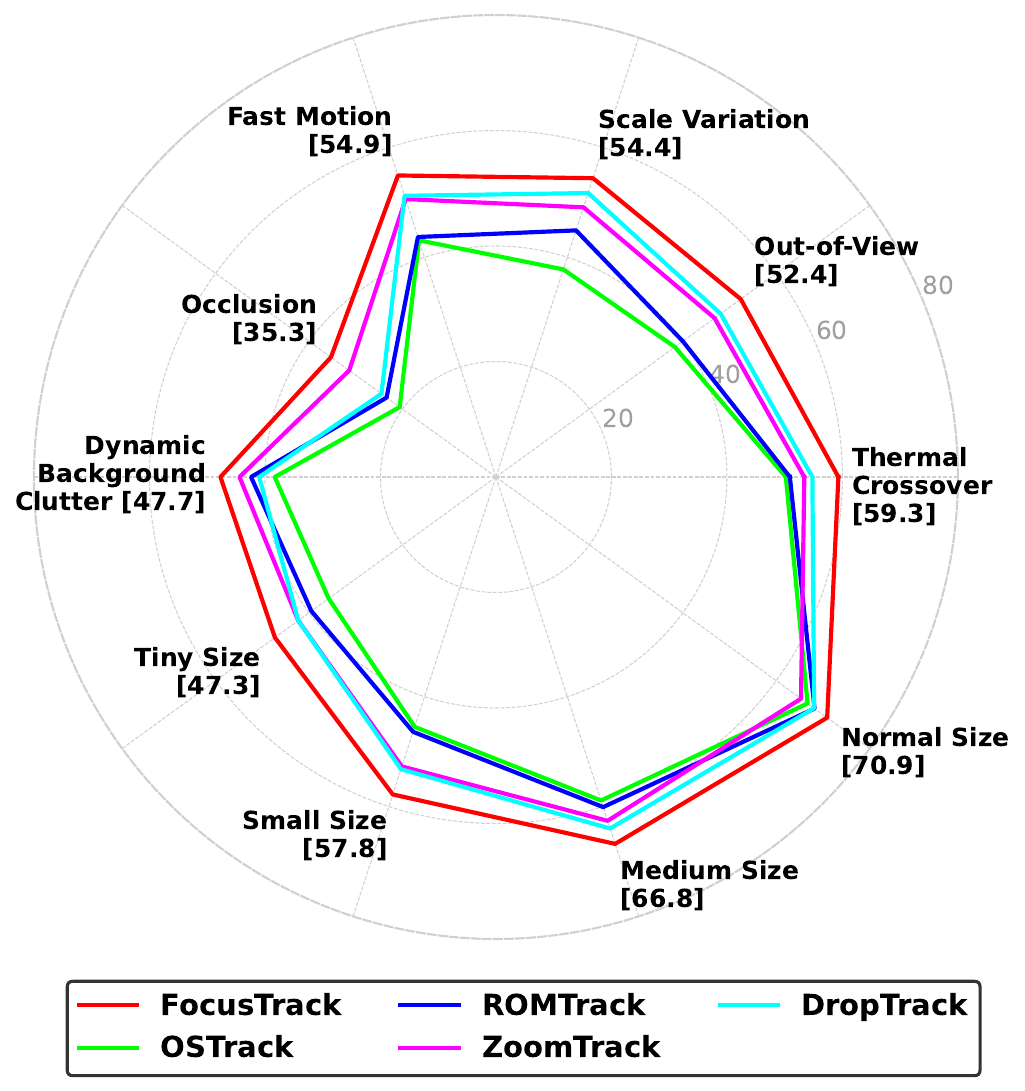}
    \caption{Attribute-based evaluation on the AntiUAV410 test set. AUC score is used to rank different trackers.}
    \label{fig:radar}
\end{figure}

\begin{figure*}[b]
\begin{center}
\vspace{-3mm}
\includegraphics[width=0.29\linewidth]{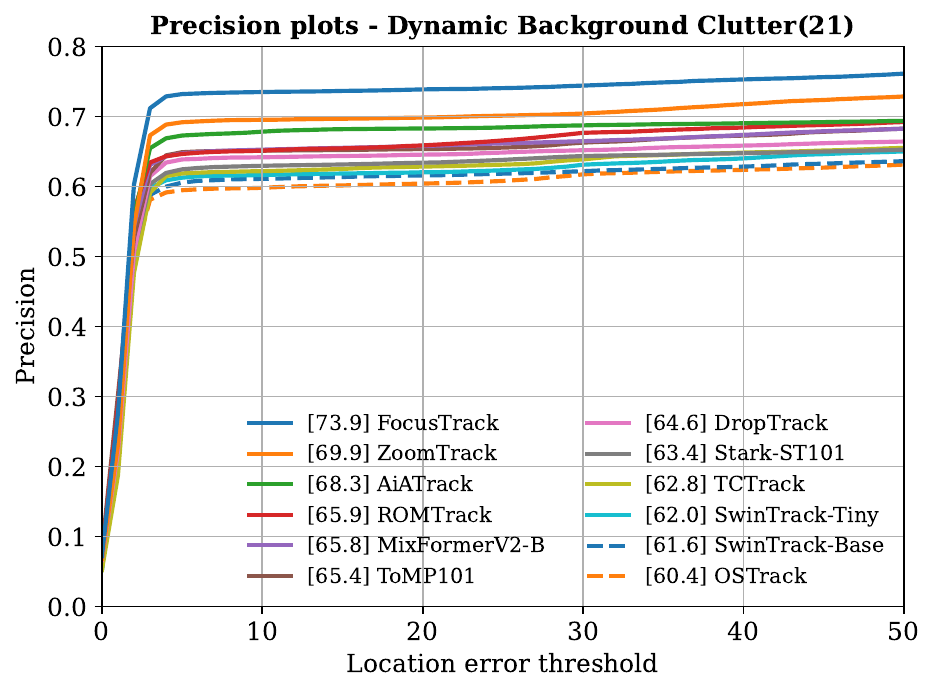}\hfill
\includegraphics[width=0.29\linewidth]{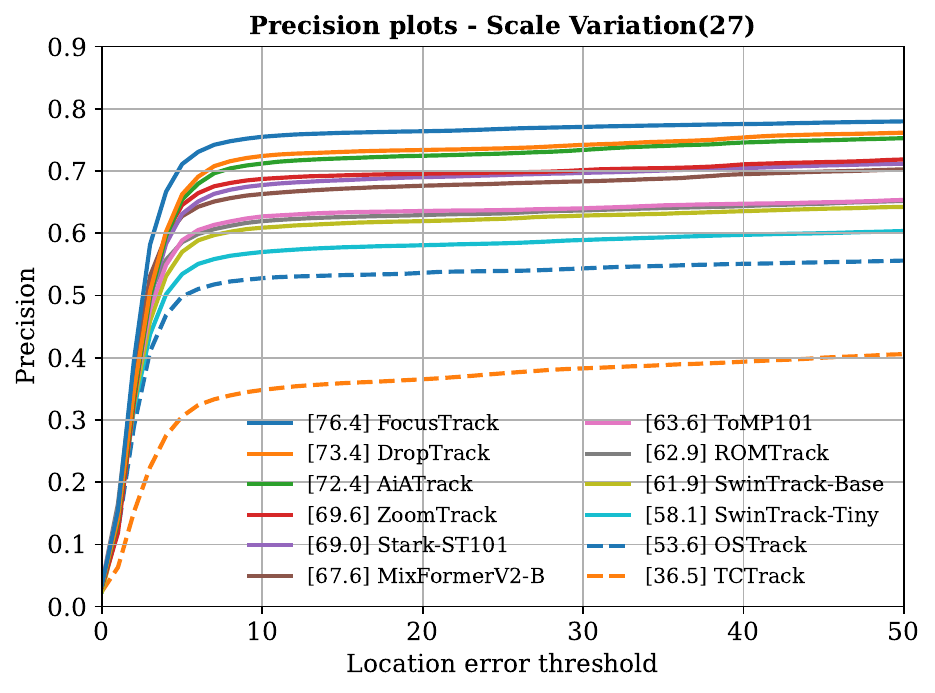}\hfill
\includegraphics[width=0.29\linewidth]{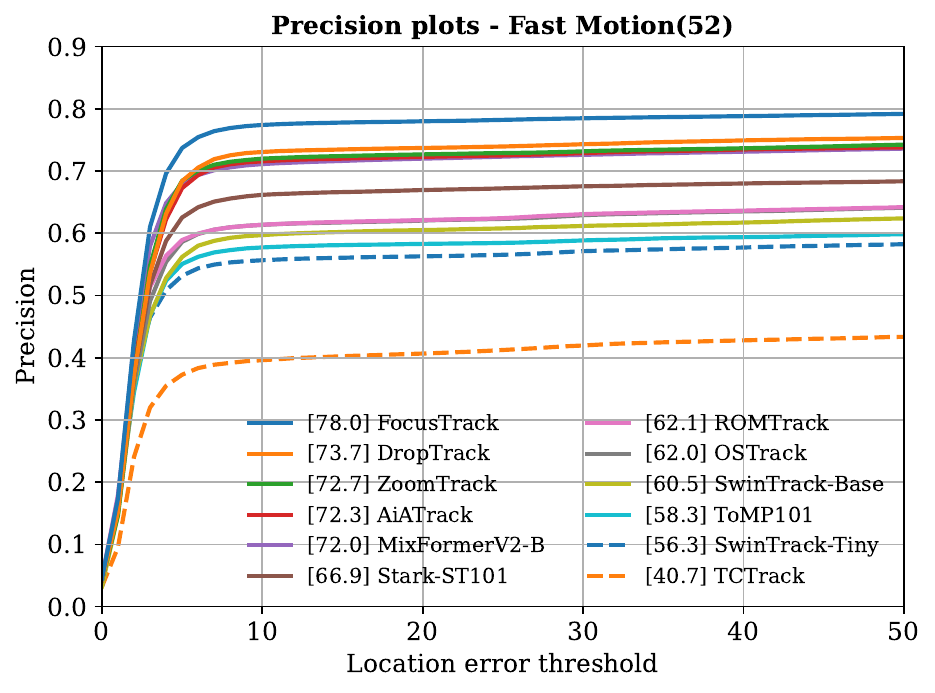}\hfill
\includegraphics[width=0.29\linewidth]{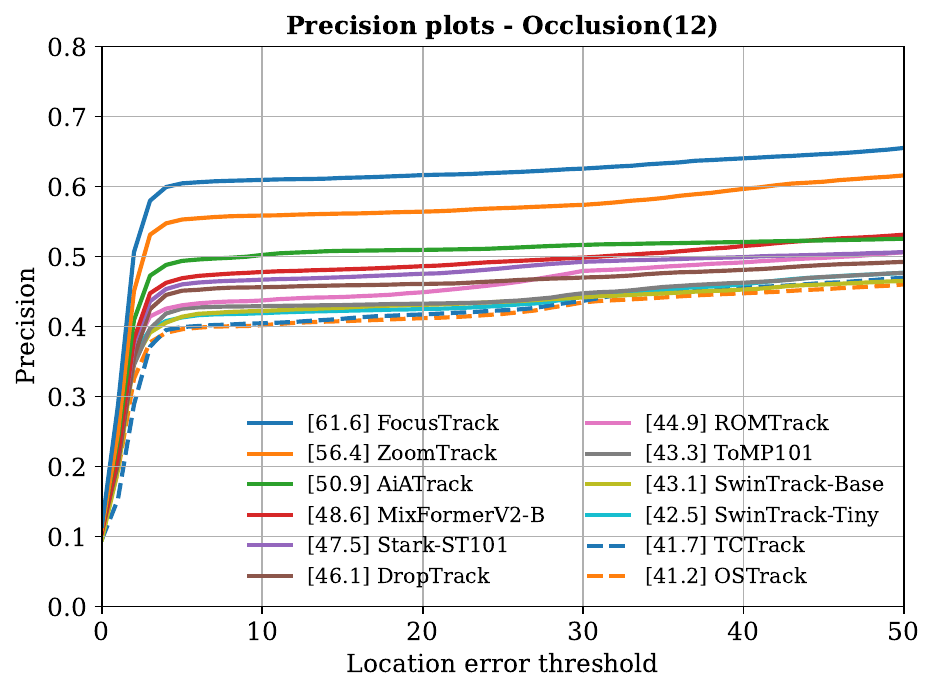}\hfill
\includegraphics[width=0.29\linewidth]{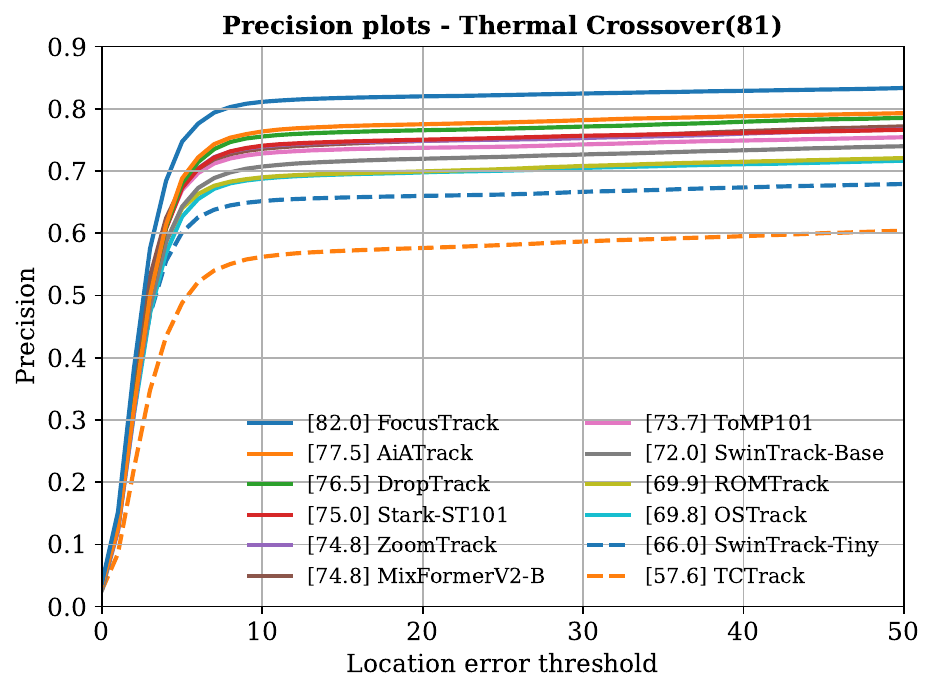}\hfill
\includegraphics[width=0.29\linewidth]{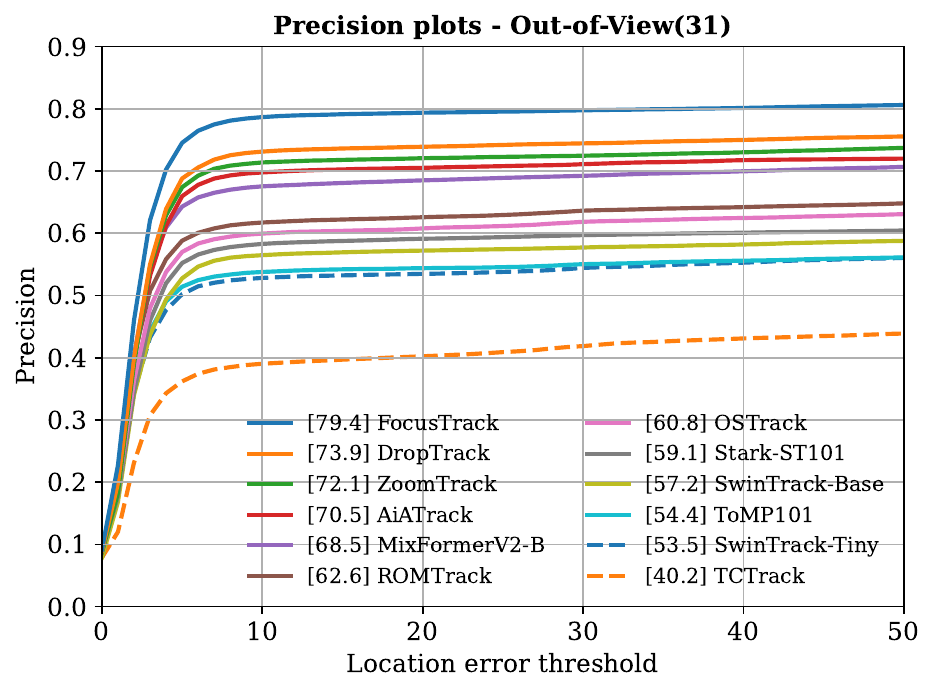}\hfill
\end{center}
\vspace{-4mm}
\caption{Attribute-based precision performance analysis on the AntiUAV410 test set. Precision plots are shown for six challenging scenarios: Dynamic Background Clutter (DBC), Scale Variation (SV), Fast Motion (FM), Occlusion (OC), Thermal Crossover (TC), and Out-of-View (OV). Better viewed in color and zoom modes.}
\vspace{-4mm}
\label{fig:attribute_precision}
\end{figure*}

\vspace{-4mm}
\subsection{Attribute-based Analysis.}

To thoroughly evaluate the tracker's robustness across diverse challenging scenarios, performance is assessed along six key attributes: Dynamic Background Clutter (DBC), Scale Variation (SV), Fast Motion (FM), Occlusion (OC), Thermal Crossover (TC), and Out-of-View (OV). As shown in Fig.~\ref{fig:attribute_success} and Fig.~\ref{fig:attribute_precision}, FocusTrack consistently outperforms other trackers across all attributes, with particularly significant advantages in challenging scenarios.

Our method demonstrates substantial improvements over the baseline (OSTrack) in handling Out-of-View (OV) scenarios, achieving gains of 14.1\% in success rate and 18.6\% in precision. This performance boost can be attributed to our SRA module's effective determination of target visibility status. For Fast Motion (FM) and Occlusion (OC) scenarios, FocusTrack exhibits robust performance with success rate improvements of 11.8\% and 14.7\%, and precision gains of 16.0\% and 19.4\%, respectively, validating its capability in handling rapid target movements and occlusion recovery.

Most notably, our tracker achieves remarkable performance in Scale Variation (SV) scenarios, with substantial improvements of 16.6\% in success rate and 22.8\% in precision. These significant gains stem from the synergistic effect of our SRA module and fine-grained hierarchical feature exploitation Module ATM, which together ensure robust feature representation across varying scales. These comprehensive results demonstrate the effectiveness of our approach in addressing diverse real-world tracking challenges.

\begin{figure*}[t]
    \centering
    \includegraphics[width=0.90\textwidth]{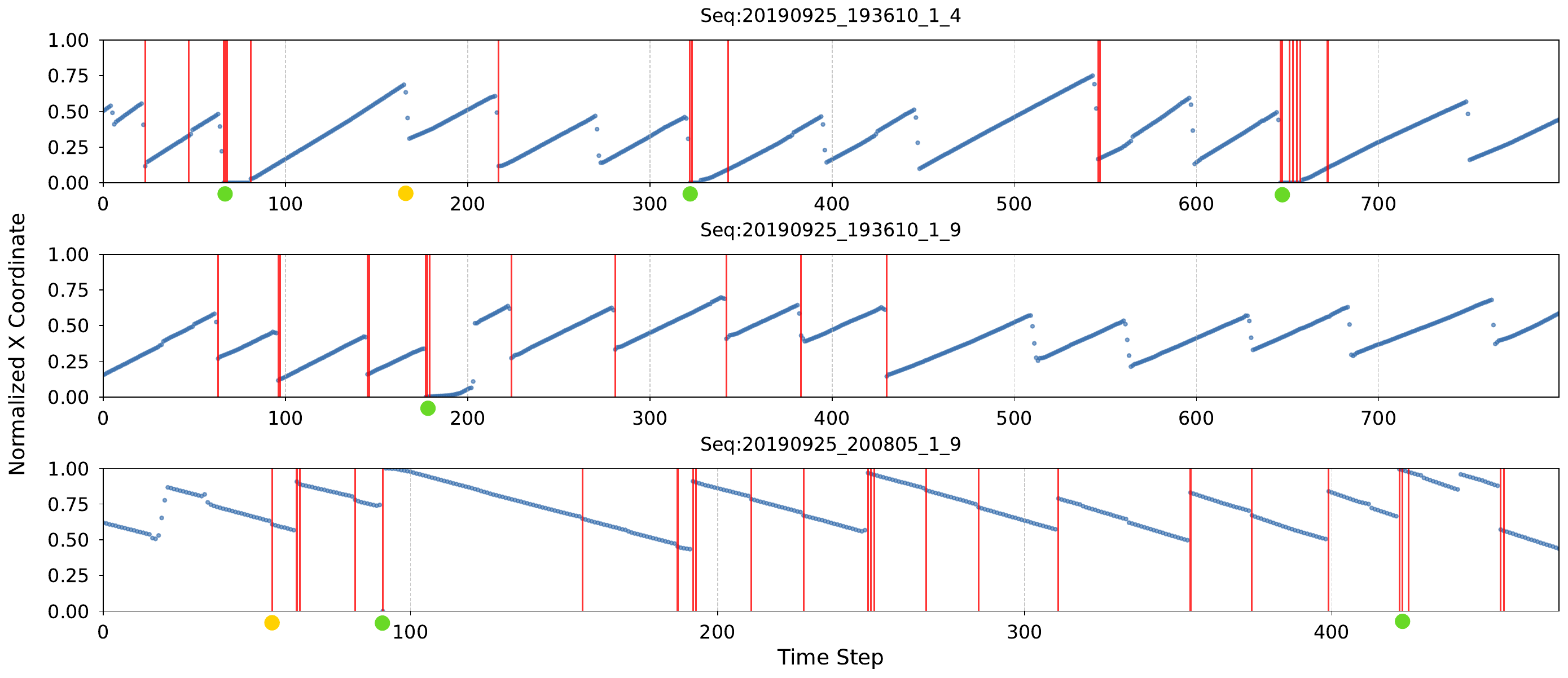}
    \caption{Qualitative analysis of SRA strategy's response to target position changes. The visualization shows target trajectories (\textcolor{blue}{blue curves}) from three representative sequences, with \textcolor{red}{red markers} indicating SRA's search region adjustments. \textcolor{new_green}{Green dots} mark frames where targets exit the field of view, while \textcolor{new_yellow}{yellow dots} highlight cases of minor target shifts (sequence 1) and Thermal Crossover-triggered adjustments (sequence 3).}
    \label{fig:sra_vis}
\end{figure*}

\begin{figure}[t]
    \centering
    \includegraphics[width=0.48\textwidth]{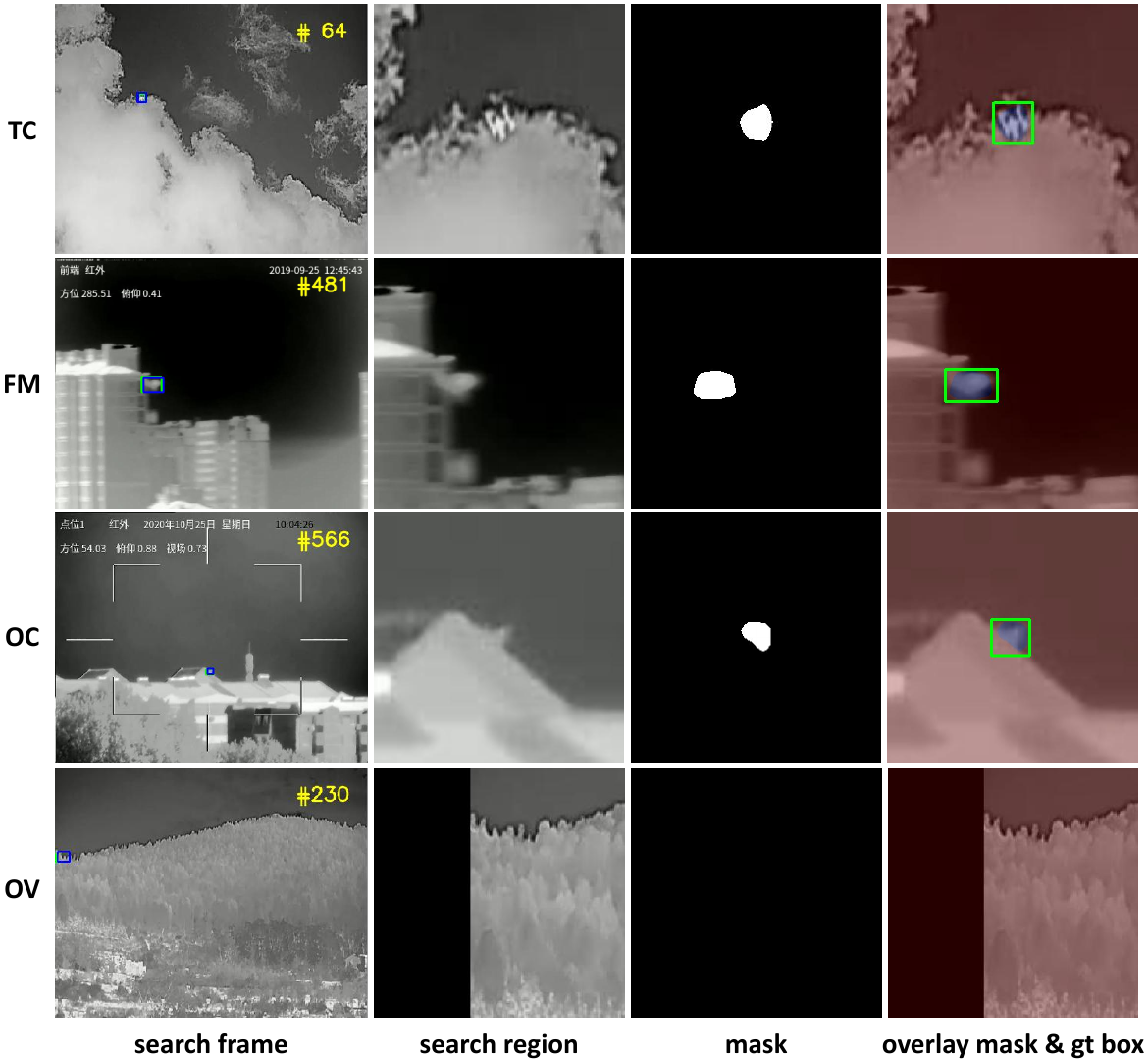}
    \caption{
    Qualitative analysis of the ATM module's segmentation performance in challenging scenarios. 
    Columns 1-4 respectively visualize the search frame, search region, predicted mask, and mask overlay with the ground truth box (green) under different attributes: Thermal Crossover (TC), Fast Motion (FM), Occlusion (OC), and Out-of-View (OV).}
    \label{fig:mask_visualization}
    \vspace{-4mm}
\end{figure}

\begin{figure*}[t]
    \centering
    \includegraphics[width=0.92\textwidth]{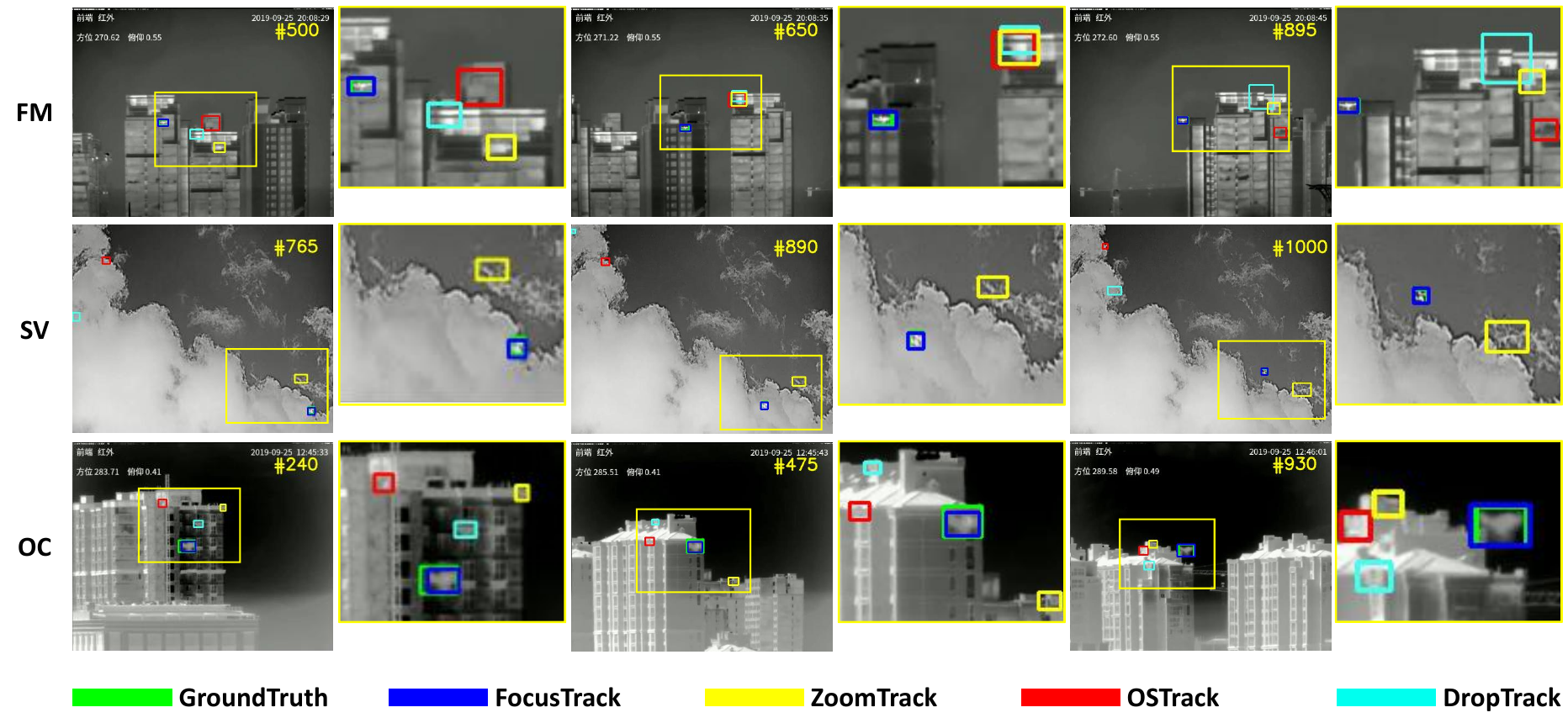}
    \vspace{-2mm}
    \caption{Qualitative visualization of four trackers under three challenging scenarios: FM (Fast Motion), SV (Scale Variation) and OC (Occlusion). Each row in the figure represents the tracking results of a video sequence with the corresponding difficulty attribute. The images in even-numbered columns are zoomed-in views of the yellow-boxed regions from the preceding odd-numbered columns. Better view in zoom.}
    \vspace{-2mm}
    \label{fig:visualization}
\end{figure*}

\begin{figure*}[h]
    \centering
    \includegraphics[width=0.92\textwidth]{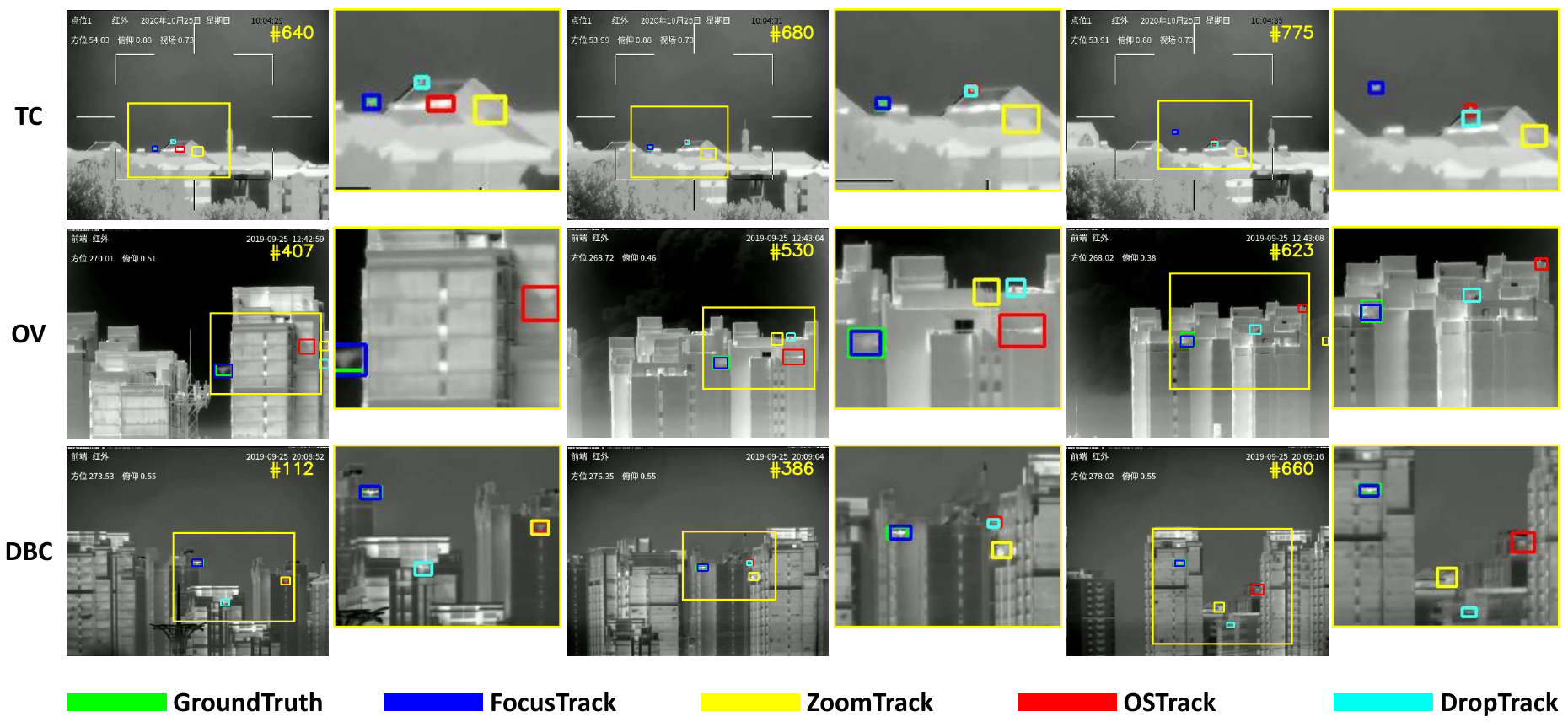}
     \vspace{-2mm}
    \caption{Qualitative visualization of four trackers under three challenging scenarios: TC (Thermal Crossover), OV (Out-of-View) and DBC (Dynamic Background Clutter). Each row in the figure represents the tracking results of a video sequence with the corresponding difficulty attribute. The images in even-numbered columns are zoomed-in views of the yellow-boxed regions from the preceding odd-numbered columns. Better view in zoom.}
    \vspace{-2mm}
    \label{fig:visualization1}
\end{figure*}

A radar plot, shown in Fig.~\ref{fig:radar}, provides a comprehensive overview of the tracker's performance across all attributes and scale variations (from Normal to Tiny Size). As shown in the plot, FocusTrack (depicted in red) consistently outperforms competing methods, forming the outermost polygon that encompasses all other trackers. 

Particularly noteworthy is our tracker's robust performance in challenging small-scale scenarios, achieving an AUC of 57.8\% for Small Size and 47.3\% for Tiny Size targets, surpassing OSTrack by 12.3\% and 11.5\%, respectively. These results further highlight the effectiveness of our method in tracking targets under extreme scale variations, ensuring stable and accurate performance across diverse conditions.

\vspace{-4mm}
\subsection{Qualitative analysis.}

A detailed qualitative analysis is performed on the SRA strategy, the ATM module, and the final tracking results, offering deeper insights into their individual contributions and overall effectiveness.

\subsubsection{Qualitative Analysis of SRA Strategy}

The effectiveness of the Search Region Adjustment (SRA) strategy is examined by visualizing its response to target position changes across three representative sequences from the AntiUAV410 test set (Fig.~\ref{fig:sra_vis}). The visualization traces normalized target x-coordinates (blue curves) over time, highlighting frames where SRA adjusts the search region (red markers).

The analysis reveals the tracker's adaptive capabilities in handling diverse target movements. 
During abrupt position shifts or instances where the target momentarily exits the field of view (indicated by green dots), the SRA module promptly modifies the search factor, facilitating rapid re-localization.

For minor target displacements, as observed in sequence 1 around frame 160 (yellow dots), the base search factor effectively sustains target visibility at the field's edge without unnecessary adjustments. This behavior confirms the SRA strategy’s position-unbiased assessment across the entire field of view.

Additionally, sequence 3 (around frame 60, yellow dots) showcases adjustments made in response to continuous target motion. Notably, these modifications stem from severe Thermal Crossover rather than actual target disappearance. Once stable conditions are restored, the search factor seamlessly reverts to its base value, ensuring efficiency without incurring unnecessary computational overhead.

\subsubsection{Qualitative Analysis of ATM Module}
The effectiveness of the Attention-to-Mask (ATM) module is examined through segmentation visualizations across various challenging scenarios from the AntiUAV410 test set (Fig.~\ref{fig:mask_visualization}).
As shown in rows 1-3, our model generates accurate segmentation masks despite challenging conditions including strong thermal interference (TC), motion blur from rapid target movement (FM), and partial target occlusion (OC). Moreover, when the target is absent from the search region in the OV case (row 4), the model correctly outputs an empty mask, demonstrating its ability to avoid false positives when the target is not present.

\subsubsection{Qualitative Visualization of Tracking Results}

A comparative visualization of tracking performance against state-of-the-art methods is provided under six challenging scenarios, as shown in Fig.~\ref{fig:visualization} and Fig.~\ref{fig:visualization1}.
The results highlight the superiority of the proposed approach in dynamically adapting the search region, escaping local traps where competing trackers fail, and effectively reacquiring the target. Even in complex environments, the tracker maintains precise localization, demonstrating remarkable resilience to visual disturbances.

\vspace{-2mm}
\section{Conclusion}
This paper presents FocusTrack, a novel tracking framework designed to dynamically adjust the search region while refining feature representations.
The proposed Search Region Adjustment (SRA) strategy dynamically modifies the field of view based on target present probability, ensuring a balance between precise localization and global search capability. 
Meanwhile, the Attention-to-Mask (ATM) module enhances target representation by integrating hierarchical features through class-specific masks. Our method achieves state-of-the-art performance on challenging Anti-UAV benchmarks, outperforming local-based trackers in accuracy while significantly improving efficiency over global search approaches. Extensive experiments prove that our approach can provide a robust and efficient solution for Anti-UAV tracking. Furthermore, with the mask output mode, our method can potentially benefit from advancements in small object detection datasets and related research\cite{st-trans}, paving the way for further performance improvements. Future work will explore leveraging segmentation priors and richer datasets to refine feature aggregation, aiming to enhance tracking robustness in extreme scenarios.

\vspace{-2mm}
\section*{Acknowledgments}
This work was financially supported by the Aeronautical Science Foundation of China (No. 2024M071072001).

\vspace{5mm}
{
\bibliographystyle{IEEEtran}
\bibliography{IEEEabrv,main}
}

\vfill

\end{document}

%% file: tables/comp_on_antiuav310.tex
\begin{table}[tp]
\caption{Comparison with state-of-the-art on the Anti-UAV test set.}

\renewcommand{\arraystretch}{1.2}
\begin{center}
\vspace{-2mm}
\resizebox{1.0\linewidth}{!}{
\begin{tabular}{c|c|c|ccc|c}
\hline
\multirow{2}{*}{Method} &
  \multirow{2}{*}{Source} &
  \multirow{2}{*}{Size} &
  \multicolumn{4}{c}{AntiUAV~\cite{antiuav310}}  \\ 
  \cline{4-7} 
    &  &  &  AUC   & P & P$_{Norm}$ & SA   \\
    \hline

OSTrack~\cite{ostrack}  & ECCV 22 & 256
& 59.2 & 79.4 & 77.5  & 60.2\\

ROMTrack~\cite{romtrack}  & ICCV 23 & 256 
& 59.4 & 78.9 & 77.1 & 60.5 \\

ZoomTrack~\cite{zoomtrack}  & NeurIPS 23 & 256 
& 63.5 & 86.0 & 83.4 & 64.5\\

DropTrack~\cite{droptrack}  & CVPR 23 & 256 
& 64.2 & 85.8 & 83.2 & 65.2\\

\hline

\textbf{FocusTrack}  & \textbf{Ours} & \textbf{256}
& \textbf{67.7} & \textbf{90.9} & \textbf{88.4} & \textbf{68.9}\\

\hline
 \end{tabular}}
\vspace{-4mm}
\end{center}
\label{tab:antiuav310}
\end{table}

%% file: tables/comp_on_antiuav410.tex
\begin{table}[t]
\caption{Comparison with state-of-the-art on the Anti-UAV410 test set. The \underline{upper} section reports results without training on the Anti-UAV410 training set, while the \underline{lower} section presents results after re-training on it.}
\renewcommand{\arraystretch}{1.2}
\begin{center}
\vspace{-2mm}
\resizebox{1\linewidth}{!}{
\begin{tabular}{c|c|c|ccc|c}
\hline
\multirow{2}{*}{Method} &
  \multirow{2}{*}{Source} &
  \multirow{2}{*}{Size} &
  \multicolumn{4}{c}{AntiUAV410~\cite{antiuav410}}  \\ 
  \cline{4-7} 
    &  &  &  AUC   & P & P$_{Norm}$ & SA   \\
    \hline

TransT~\cite{transt}  & CVPR 21 & 256 
& 48.2 & 67.7 & 64.1 & 48.9\\

ETTrack~\cite{ettrack}  & WACV 23 & 256 
& 41.5 & 59.7  & 54.8   & 41.6\\

MixFormerV2-S~\cite{mixformerv2}  & NeurIPS 23 & 224 
& 45.6 & 64.1  & 60.0  & 46.1\\

GRM~\cite{grm}  & CVPR 23 & 256
& 42.3 & 58.5 & 55.1 & 42.2\\

ARTrack~\cite{artrack}  & CVPR 23 & 256
& 48.2 & 67.2 & 62.9 & 48.5\\

JointNLT~\cite{jointnlt}  & CVPR 23 & 320 
& 48.4 & 69.0 & 64.5 & 48.9\\

SeqTrack~\cite{seqtrack}  & CVPR 23 & 256
& 52.2 & 73.8 & 70.0 & 52.9\\

PromptVT~\cite{promptvt}  & TCSVT 24 & 320 
& 50.5 & 71.5 & 65.6 & 51.2\\

\midrule
\midrule

Stark-ST101~\cite{stark}  & ICCV 21 & 320 
& 56.2 & 78.5 & 74.6 & 57.1\\

TCTrack~\cite{tctrack}  & CVPR 22 & 287
& 41.1 & 60.4  & 56.0  & 41.6  \\

% % Transformer based
% HCAT~\cite{hcat}  & ECCV 22 & 256 & 48.9
% & 44.8 & 60.6  & 59.7 \\

OSTrack~\cite{ostrack}  & ECCV 22 & 256
& 53.7 & 73.9 & 70.9  & 54.7\\

ToMP50~\cite{tomp}  & CVPR 22 & 288
& 54.1 & 73.8 & 70.2  & 55.1\\

ToMP101~\cite{tomp}  & CVPR 22 & 288 
& 54.2 & 75 & 70.5 & 55.1\\

SwinTrack-Tiny~\cite{swintrack}  & NeurIPS 22 & 224 
& 53.0 & 71.4 & 68.1 & 53.1\\

SwinTrack-Base~\cite{swintrack}  & NeurIPS 22 & 384 
& 55.9 & 76.4 & 72.3 & 55.7\\

AiATrack~\cite{aiatrack}  & ECCV 22 & 320 
& 58.6 & 82.3 & 78.0 & 59.6\\

ROMTrack~\cite{romtrack}  & ICCV 23 & 256 
& 54.7 & 74.5 & 71.7 & 55.7 \\

ZoomTrack~\cite{zoomtrack}  & NeurIPS 23 & 256 
& 58.4 & 81.2 & 77.4 & 59.4\\

MixFormerV2-B~\cite{mixformerv2}  & NeurIPS 23 & 288 
& 58.7 & 80.5 & 76.8 & 59.6\\

DropTrack~\cite{droptrack}  & CVPR 23 & 256
& 59.2 & 82.2 & 78.2 & 60.2\\

\hline

\textbf{FocusTrack}  & \textbf{Ours} & \textbf{256}
& \textbf{62.8} & \textbf{86.2} & \textbf{82.8} & \textbf{63.9}\\

\hline
 \end{tabular}}
\end{center}
\vspace{-4mm}
\label{tab:antiuav410}
\end{table}

%% file: tables/fps.tex
\begin{table}[tp]
    \centering
    \caption{Comparison of Spees(fps) and MACs(G). These test results were obtained on the same machine.}
    \vspace{-2mm}
    \renewcommand{\arraystretch}{1.1}
    \resizebox{1.0\linewidth}{!}
    {
    \begin{tabular}{c|c|cc|cc|c}
     \toprule

\multirow{2}*{Type} & \multirow{2}*{Tracker} & Speed  & MACs & \multirow{2}*{AUC} 
& \multirow{2}*{P} & \multirow{2}*{SA} \\ 
    && (fps$\uparrow$)    &  (G$\downarrow$)  & &  &  \\

\midrule

\multirow{6}*{Local} &
     
     OSTrack~\cite{ostrack}& 137  & 29.1 
     & 53.7 & 73.9  & 54.7\\
     
     &ROMTrack~\cite{romtrack}& 8  & 34.5
     & 54.7 & 74.5 & 55.7 \\

     &ZoomTrack~\cite{zoomtrack}& 154  & 29.1
     & 58.4 & 81.2 & 59.4\\

     &DropTrack~\cite{droptrack}& 151  & 29.1
     & 59.2 & 82.2  & 60.2\\
    
     % \hline
     &FocusTrack(SRA) & 143  & 29.1
     & 62.3 & 84.1 & 63.2\\

     &FocusTrack(Ours) & 44 & 30.1
     & 62.8 & 86.2 & 63.9\\

    \midrule

     Global & SiamDT~\cite{antiuav410}  & 8   & 225.3
     & 66.8 & 90.0  & 68.2\\
     
     \bottomrule

    \end{tabular}
    }
    \vspace{-4mm}
    \label{tab:efficiency}
\end{table}

%% file: tables/ablation.tex
\begin{table}[ht]
    \centering
    % \fontsize{7}{8}\selectfont\
    \setlength{\tabcolsep}{4pt} 
    \caption{Ablation Study of different components in FocusTrack on AntiUAV410 test set.}
    \vspace{-2mm}
    \begin{tabular}{c|ccllc|c|c}
    \toprule
    \multirow{2}*{No.} &\multirow{2}*{sf=6} & DropMAE  & \multirow{2}*{SRA} & \multirow{2}*{ATM}  & Two-phase & \multirow{2}*{AUC} & \multirow{2}*{$\Delta$}\\
    &&Pretrained&&&  Training &&\\

    \midrule
    1 &     &   &   &   &   & 53.7 & -  \\
    2 &  \checkmark   &   &   &   &   & 59.8 & +6.1  \\
    3 &   \checkmark  & \checkmark  &   &   &   & 60.2 & +6.5 \\
    4 & \checkmark    & \checkmark  & \checkmark  &   &   & 62.3 & +8.6 \\
    5 &  \checkmark   & \checkmark  &   & \checkmark  &   & 60.6 & +6.9 \\
    6 & \checkmark    & \checkmark &  \checkmark &  \checkmark &   & 62.6 & +8.9 \\
    
    7 &  \checkmark  &  \checkmark &  \checkmark & \checkmark  & \checkmark  & \textbf{62.8} & \textbf{+9.1} \\

    \bottomrule
    \end{tabular}
    \label{tab:ablation}
    \vspace{-3mm}
\end{table}

%% file: tables/ablation_sra_training.tex
\begin{table}[tp]
\centering
\caption{Ablation Study on Search Region Adjustment Module in training phase.}
\vspace{-2mm}
\begin{tabular}{c|c|c|ccc}
\toprule
No. & Pooling & MLP Layers &  AUC  &  P & $\rm P_{norm}$ \\

\midrule 
1 & -                   &  2    & 61.6 & 85.1 & 81.4\\
2 & Average Pooling     &  2    & 61.7 & 85.4 & 81.7\\  
3 & Attention Pooling   &  2    & 62.6 & 86.7 & 83.0 \\
4 & Attention Pooling   &  1    & 60.4 & 83.5 & 79.9\\
\bottomrule
\end{tabular}
\vspace{-4mm}
\label{tab:sra_train}
\end{table}

%% file: tables/ablation_sra_inference.tex
\begin{table*}[t]
    \centering
    \caption{Ablation Study on Search Region Adjustment Module in inference phase.}

    \resizebox{1\textwidth}{!}{
    % \fontsize{6}{10}\selectfont
    \begin{tabular}{ccccc|cccc|cccc|cccc}
     \toprule
     \multicolumn{5}{c|}{(a) Study on \( T_{\text{logits}} \)}  &\multicolumn{4}{c|}{(b) Study on \( T_{\text{score}} \)} &\multicolumn{4}{c|}{(c) Study on \( f^{\text{max}}_{x} \)}
     &\multicolumn{4}{c}{(d) Study on \( f^{\text{step}}_{x} \)}
     \\
     \midrule
     
      \multicolumn{1}{c|}{No.} 
      & \multicolumn{1}{c|}{\( T_{\text{logits}} \)} 
      &\multicolumn{1}{c}{AUC} 
      &\multicolumn{1}{c}{P} 
      &\multicolumn{1}{c|}{$\rm P_{norm}$} 
      &\multicolumn{1}{c|}{\( T_{\text{score}} \)}
      &\multicolumn{1}{c}{AUC} 
      &\multicolumn{1}{c}{P} 
      &\multicolumn{1}{c|}{$\rm P_{norm}$} 
      &\multicolumn{1}{c|}{\( f^{\text{max}}_{x} \)}
      &\multicolumn{1}{c}{AUC} 
      &\multicolumn{1}{c}{P} 
      &\multicolumn{1}{c|}{$\rm P_{norm}$} 
      &\multicolumn{1}{c|}{\( f^{\text{step}}_{x} \)}
      &\multicolumn{1}{c}{AUC} 
      &\multicolumn{1}{c}{P} 
      &\multicolumn{1}{c}{$\rm P_{norm}$} 
      \\
      \midrule
      
      \multicolumn{1}{c|}{1} 
      &\multicolumn{1}{c|}{0.7} &61.7 & 85.5 & 81.9 
      &\multicolumn{1}{c|}{0.4} &61.5 & 85.2 & 81.6 
      &\multicolumn{1}{c|}{8}   &\textbf{62.6} & \textbf{86.7} & \textbf{83.0}
      &\multicolumn{1}{c|}{0.5} &62.0 & 85.9 & 82.3 \\

      \multicolumn{1}{c|}{2} 
      &\multicolumn{1}{c|}{0.8} &\textbf{62.6} & \textbf{86.7} & \textbf{83.0}
      &\multicolumn{1}{c|}{0.5} &\textbf{62.6} & \textbf{86.7} & \textbf{83.0}
      &\multicolumn{1}{c|}{9} &61.9 & 85.7 & 82.0 
      &\multicolumn{1}{c|}{0.75} &62.4 & 86.4 & 82.8  \\

      \multicolumn{1}{c|}{3} 
      &\multicolumn{1}{c|}{0.9} &62.5 & 86.6 & 82.9
      &\multicolumn{1}{c|}{0.6} &62.3 & 86.2 & 82.6 
      &\multicolumn{1}{c|}{10} &61.1 & 84.6 & 80.8 
      &\multicolumn{1}{c|}{1} &\textbf{62.6} & \textbf{86.7} & \textbf{83.0} \\
    \bottomrule
     
    \end{tabular}
    }
\vspace{-4mm}
\label{tab:sra_inference}
\end{table*}

%% file: tables/ablation_atm.tex
\begin{table}[tp]
\centering
\setlength{\tabcolsep}{8pt}
\caption{Ablation Studies on Attention-to-Mask Module.}
\begin{tabular}{c|c|ccc|c}
\toprule
Num Layers & Used Layers & AUC &  P & $\rm P_{norm}$ & SA  \\
\midrule 
1 & [12]          & 61.8 &  85.5 & 81.8 & 62.8\\
2 & [6, 12]       &  62.2 &  85.9 & 82.3 & 63.2 \\  
3 & \textbf{[6, 8, 12]}    & \textbf{62.6} &  \textbf{86.7} & \textbf{83.0} & \textbf{63.6}\\
3 & [4, 6, 12]    & 62.3  &  86.2 & 82.3 & 63.4\\
3 & [4, 8, 12]    & 62.1  &  86.0 & 82.1 & 63.1\\
4 & [3, 6, 9, 12] & 61.3  &  84.8 & 81.0 &  62.3\\ 
\bottomrule
\end{tabular}
\vspace{-5mm}
\label{tab:atm}
\end{table}

%% file: tables/loss_function.tex
\begin{table}[tp]
\centering
\vspace{-2mm}
\caption{Ablation study on loss functions for ATM module, evaluated on the AntiUAV410 test set.}
\vspace{-2mm}
\setlength{\tabcolsep}{11pt}
\label{tab:loss_ablation}
\begin{tabular}{c|ccc|c}
\toprule
Loss Function  &  AUC  &  P & $\rm P_{norm}$ & SA \\

\midrule 
w/o supervision     &  61.0     & 82.3  &  80.8  & 62.0 \\
Dice Loss           &  61.5     & 85.0  &  81.3  & 62.5 \\  
\textbf{Focal Loss} &  \textbf{62.6} & \textbf{84.6}  &  \textbf{83.0}  & \textbf{63.6} \\
Dice \& Focal Loss  &  62.0     & 83.8  &  82.1  & 63.0 \\
\bottomrule
\end{tabular}
\vspace{-4mm}
\end{table}